\documentclass{article}

\usepackage[english]{babel}
\usepackage[letterpaper,top=2cm,bottom=2cm,left=3cm,right=3cm,marginparwidth=1.75cm]{geometry}

\usepackage{amsmath}
\usepackage{graphicx}
\usepackage[colorlinks=true, allcolors=blue]{hyperref}
\usepackage{comment}
\usepackage{booktabs}
\usepackage{float}
\usepackage{xurl}
\usepackage[round,authoryear]{natbib}
\usepackage{endnotes}
\let\footnote=\endnote

\title{Automated sign detection across the Electronic Babylonian Library: A large-scale dataset and end-to-end cuneiform OCR pipeline}

\author{
Wentao Che$^{1}$,
Esteban Garcés Arias$^{2,3}$,
Asim Niaz$^{1}$,\\
Andreas Bender$^{2,3}$,
Enrique Jiménez$^{1}$\\[0.8em]
$^{1}$Institute of Assyriology and Hittite Studies, LMU Munich, Germany\\
$^{2}$Department of Statistics, LMU Munich, Germany\\
$^{3}$Munich Center for Machine Learning (MCML), Germany
}

\date{}

\begin{document}
\maketitle

\begin{abstract}
Learning to read cuneiform tablets is an extremely demanding task; consequently, of the roughly half million excavated tablets, only a small fraction has been analysed by Assyriologists. Computer vision offers a promising avenue for decipherment but requires large, densely annotated datasets. To address this limitation, the largest annotated cuneiform sign dataset to date is used, and a Deformable Detection Transformer (DETR)-based object detection model is evaluated under two class granularities of 173 and 106 classes. The proposed system integrates automatic tablet-side extraction, heuristic line grouping, and n-gram--based textual similarity evaluation to bridge visual sign detection and textual structure, and achieves consistent improvements of up to 28--37\% over prior work on COCO-style detection metrics.
At inference, the method is applied to 87,668 tablet fragments from the Electronic Babylonian Library (eBL) corpus, producing nearly 2.9 million sign detections. Although the approach operates without linguistic priors and remains sensitive to tablet damage and layout variability, it provides a scalable and interpretable foundation for corpus-wide cuneiform analysis and supports future integration with multimodal and linguistic modelling frameworks.
\end{abstract}

\section{Introduction}

Cuneiform is one of the oldest known writing systems, spanning more than three millennia and utilised for numerous languages of the Ancient Near East. Its geographic spread reflects its usage in administrative, legal, economic, and literary texts. Unlike modern handwriting on paper, cuneiform was written by pressing a reed stylus into wet clay to create wedge-shaped impressions \citep{saeed2024create, Yesiltepe2019}. Although at least 500,000 tablets are preserved in museums worldwide \citep{Streck2010}, most remain unread or only partially studied. The main reason is that the scale of the corpus far exceeds the ability of a limited number of trained Assyriologists available to analyse it. Furthermore, thousands of years of environmental exposure resulted in erosion, cracks, and missing wedges on many cuneiform tablets, which introduced substantial noise and ambiguity \citep{reade2017manufacture}. This fact complicates both manual interpretation and automated recognition. With the disappearance of the ancient scribal tradition, modern scholarship depends on reconstructing meaning from incomplete corpora, multilingual parallels, and philological inference. Together, these factors define the methodological challenges of the study of cuneiform.

Deciphering cuneiform tablets manually is a formidable task that only a handful of expert Assyriologists worldwide can undertake. Even the initial step of sign identification is difficult due to the vast inventory of cuneiform signs and the deteriorated state of many tablets. Digitisation and computational tools thus increasingly accelerate Assyriological research by allowing large-scale search, comparison, and reconstruction of fragmentary texts. Machine-learning approaches in computer vision, especially convolutional neural networks (CNNs) and Vision Transformer (ViT) architectures, can assist in sign detection and classification, reducing the burden on experts. In this context, object detection (OD) localises and classifies individual signs within tablet images, while optical character recognition (OCR) refers to the broader task of converting these visual detections into structured, machine-readable form.

Even though OD and OCR have improved considerably for modern and historical texts, cuneiform is still far more challenging to handle than typical two-dimensional scripts. First, the script itself is highly complex, as sign shapes, orientations, and spatial arrangements vary considerably across time periods and scribal traditions \citep{bogacz2022digital}. Small geometric differences can distinguish one sign from another, making sign recognition a fine-grained visual classification problem. Existing OCR models designed for alphabetic scripts are not suitable for cuneiform \citep{dencker2020deep}, as the wedge-shaped impressions are three-dimensional and fundamentally different from line-based characters \citep{gordin2024cured}. 

Beyond these script-specific difficulties, modern deep-learning approaches for detection and recognition are data-hungry: achieving competitive performance typically requires large, densely annotated training corpora \citep{dencker2020deep, williams2025deepscribe}. For cuneiform, however, producing such annotations demands specialist philological expertise, and the labelling effort scales poorly with corpus size. As a result, fully supervised datasets have remained scarce, constituting a primary bottleneck for progress in cuneiform OCR.

Addressing these challenges is essential for large-scale digitisation, analysis, and long-term preservation of ancient Mesopotamian cultural heritage. 
In this study, we build on recent work in cuneiform sign detection by focusing on the reproducibility and extensibility of the DETR-based model introduced by \citet{Cobanoglu2024}, who developed an open-source cuneiform dataset on the Electronic Babylonian Library (eBL) platform \citep{eBLPlatform}, and evaluated two object detection models. To that end, we utilise the recently published large-scale dataset of annotated cuneiform sign images \citep{lewenstein2026large}. 
Our goal is to evaluate its applicability to larger and more diverse corpora, and to adapt the workflow to the structure of the eBL dataset. This work makes the following contributions:

\begin{itemize}
    \item We provide a reproducible and extensible re-implementation of the DETR-based cuneiform sign detection model introduced by \citet{Cobanoglu2024}.
    \item We utilise an expanded version of the original dataset, growing it from 52,102 to 124,504 individually annotated signs \citep{lewenstein2026large}. It constitutes the largest publicly available supervised dataset for cuneiform sign detection to date.
    \item We adapt the training and inference workflow to the structure of the eBL dataset and introduce additional inference-time tools designed to improve the usability of raw model outputs.
    \item We conduct comprehensive evaluations that highlight the benefits and limitations of DETR-based approaches for cuneiform OD and OCR tasks.
\end{itemize}

\section{Related Work}

We organise prior work into three thematic areas:
(i)~handcrafted and classical methods for cuneiform sign recognition,
(ii)~deep-learning approaches for cuneiform detection and OCR, and
(iii)~transformer-based architectures for OCR and handwritten text recognition (HTR) in broader
historical document analysis, which provides the methodological
context for our architectural choice.

\subsection*{Handcrafted and classical methods}

In computer vision, handwriting and text recognition are accomplished
through three primary methods: character-based
\citep{wang2011end, wang2012end}, word-based
\citep{liu2017deep, zhou2017east}, and line-based approaches
\citep{wigington2018start, chammas2018handwriting}.
Unlike character-based techniques, which identify and classify a sign
in a single step, word-based and line-based techniques locate whole
words or lines without any explicit localisation at the sign level.
The latter two approaches are ill-suited to cuneiform text detection:
line-based methods can be disrupted by the damaged or irregular lines
that are common in cuneiform fragments, while the polyvalent and often
ambiguous nature of cuneiform signs prevents reliable word-level
decomposition.
This research therefore employs a character-based strategy that
enables the detector to produce bounding boxes for each cuneiform sign
directly, mirroring the sign-by-sign workflow of traditional epigraphy
while enabling scalable computational analysis.

Ahmed~et~al.\ introduced a cuneiform detection method that enhances
the tablet image, skeletonises each symbol, extracts wedge lines, and
converts these features into a Symbol Structural Vector (SSV) matched
against a stored database~\citep{ahmed2012online}.
This approach works well when symbols are clean and clearly drawn.
However, the system is sensitive to noise, distortions, and incomplete
wedges, and its accuracy decreases when symbols vary in style or
preservation.
Because the SSV relies on simple geometric features, it cannot
capture the full range of variation found in real tablets, limiting
its ability to generalise to damaged or irregular signs.

Aktas~et~al.\ applied computer-based image processing and pattern
recognition techniques to identify Hittite cuneiform signs
\citep{Yesiltepe2019}, using algorithms such as SIFT, SURF, ORB,
Hausdorff distance, and geometric feature extraction to match signs on
tablets with high-resolution reference images.
The work also includes data-mining methods to group similar signs and
an expert-system approach for rule-based translation using basic
Hittite grammar.
While the method demonstrates that automated sign recognition and
preliminary translation are technically feasible, its performance is
limited by noise, damaged or incomplete signs, and variation in
writing style.
Its dependence on handcrafted features and fixed grammar rules reduces
reliability when signs or linguistic structures become more complex.

Both systems share a common limitation: their feature spaces are
designed for idealised sign forms and do not accommodate the
diachronic variation and physical degradation typical of real tablet
collections.

\subsection*{Deep-learning approaches for cuneiform detection and OCR}

While these early systems relied heavily on handcrafted features and
explicit shape matching, later work shifted toward machine-learning
models that reduce manual feature design and handle greater visual
variability.
Research on automated cuneiform analysis spans both visual and
linguistic domains.

Early OCR-oriented work relied on weak supervision through
transliteration alignment, with \citet{dencker2020deep} demonstrating that sign detectors can be
trained without explicit bounding-box annotations.
However, their reliance on transliterations introduces noisy
pseudo-bounding boxes, and alignment errors may accumulate over the
iterative training process.
More recent systems instead employ fully supervised deep-learning
architectures such as the Single Shot MultiBox Detector (SSD)~\citep{liu2016ssd},
RetinaNet~\citep{williams2025deepscribe}, or
FCENet~\citep{zhu2021fourier} for sign detection.

DeepScribe presents a modular, vision-based method for Elamite
cuneiform OCR, using a RetinaNet detector and a ResNet classifier to
localise and identify signs from the Persepolis Fortification Archive,
trained on approximately 100{,}000 annotated
signs~\citep{williams2025deepscribe}.
It demonstrates that contemporary object-detection and classification
models can provide meaningful sign-level suggestions to scholars, even
though full transliteration requires linguistic modelling that goes
beyond image-based analysis.
A key limitation is that its vision-only framework does not yet yield
dependable end-to-end transliterations: performance declines when
classifications rely on automatically detected regions, resulting in
elevated error rates, and the system remains challenged by infrequent
signs, damaged or poorly lit impressions, and instances where
contextual linguistic information is essential.
This limitation of decoupled two-stage pipelines directly motivates
our adoption of an end-to-end detection framework.

Although significant progress has been made in cuneiform detection,
data size, annotation schemes, and high variability in the visual
representations of tablets remain a challenge.
One of the most comprehensive benchmark datasets has been offered
by~\citet{Cobanoglu2024}, confirming that a simpler end-to-end model
such as Deformable DETR can achieve better performance compared to
more complex pipelines and achieve faster inference.
Our work builds directly on~\citet{Cobanoglu2024}, more than doubling
the benchmark dataset from 52{,}102 to 124{,}504 annotations, and
extending the inference pipeline with automatic tablet-side
extraction, density-based spatial clustering of applications with noise (DBSCAN)-based line grouping, and n-gram textual evaluation.

Recent work has begun to model the internal structure of cuneiform
signs rather than treating them as purely categorical.
ProtoSnap proposes an unsupervised method that aligns skeleton-based
prototype representations to photographed cuneiform signs using deep
image features and structural priors, enabling the recovery of
fine-grained palaeographic variation~\citep{mikulinsky2025protosnap}.
While ProtoSnap focuses on internal sign morphology and operates on
isolated sign images, our work addresses the complementary problem of
large-scale sign detection from complete tablet images.

\subsection*{Transformer architectures for OCR and HTR}

The broader OCR and HTR community has undergone a decisive
architectural shift from convolutional neural network--recurrent neural network (CNN--RNN) pipelines toward transformer-based
end-to-end models, and this shift directly informs our architectural
choices.
The original DETR~\citep{carion2020endtoendobjectdetectiontransformers},
introduced by Carion~et~al., framed object detection as a direct
set-prediction problem using bipartite matching and a transformer
encoder-decoder, eliminating the need for non-maximum suppression and
hand-designed anchor generation.
Deformable
DETR~\citep{zhu2021deformabledetrdeformabletransformers} resolved
DETR's slow convergence and limited multi-scale feature resolution by
restricting attention to a small set of key sampling points around
each reference, achieving comparable accuracy with a fraction of the
training epochs.
This is a critical practical advantage, given the high-resolution
tablet images and GPU memory constraints of our training setup.

In the domain of text recognition,
TrOCR~\citep{li2022trocrtransformerbasedopticalcharacter} demonstrated
that a pure encoder-decoder transformer, pre-trained at a large scale
and fine-tuned on printed and handwritten benchmarks, substantially
outperforms CNN-RNN approaches on standard OCR tasks.
Its success illustrates the generality of the encoder-decoder paradigm
for text recognition, though cuneiform presents additional challenges
absent from standard HTR benchmarks: the three-dimensional nature of
wedge impressions, fine-grained visual similarity between sign
classes, and the absence of conventional lexical context.

For historical handwritten text recognition at the infrastructure
level, Transkribus~\citep{8270253} has become a dominant platform for
transcript production from historical manuscripts, deployed across
archives and libraries worldwide. 
It integrates layout analysis, line segmentation, and text recognition in a single pipeline---a design principle reflected in our tablet-side extraction and
line-grouping components.

The applicability of transformer-based HTR to low-resource and ancient
scripts has been confirmed by a growing body of recent work.
Garces~Arias~et~al.\ showed that a Swin image encoder~\citep{liu2021swintransformerhierarchicalvision} paired with a
BERT-based decoder~\citep{devlin2019bertpretrainingdeepbidirectional}, augmented with synthetic training data, yields
state-of-the-art HTR for Old Occitan, outperforming fine-tuned TrOCR
and commercial systems~\citep{arias-etal-2023-automatic}.
Koch~et~al.\ developed a tailored end-to-end pipeline for Medieval
Latin dictionary cards, achieving a character error rate of 1.5\% via
a vision encoder-GPT-2 decoder architecture ~\citep{Radford2019LanguageMA} with extensive data
augmentation~\citep{koch-etal-2023-tailored}.
Pavlopoulos~et~al.\ introduced a shared task on error correction in
HTR output for Byzantine Greek manuscripts and papyri spanning seven
centuries, demonstrating that post-OCR correction remains an open
challenge even for strong sequence
models~\citep{pavlopoulos-etal-2024-challenging}.
Sarawgi~et~al.\ presented a comprehensive HTR pipeline for Old Nepali
manuscripts, establishing the feasibility of large-scale digitisation
of non-Latin historical scripts under low-resource
conditions~\citep{sarawgi2025digitizingnepalswrittenheritage}.
Collectively, this body of work indicates that transformer-based
architectures generalise across scripts, time periods, and resource
levels, and underscores that cuneiform, with its unique
three-dimensional character and extreme class imbalance, represents
an open frontier for these methods.

The three specific gaps our work closes relative to prior
cuneiform-specific research are:
(1)~\textit{dataset scale}: usage of the largest supervised annotation set
for cuneiform sign detection to date \citep{lewenstein2026large};
(2)~\textit{inference tooling}: automated tablet-side extraction
and scale-invariant line grouping absent from prior systems; and
(3)~\textit{corpus-wide evaluation}: large-scale inference and
n-gram textual assessment across 87{,}668 fragments of the eBL corpus.

\section{Dataset}

The dataset comprises photographs of cuneiform tablets from a wide range of museum and archival collections, each contributing distinct imaging conditions, preservation states, and historical contexts.

\subsection{Source and Annotation}

\begin{figure}[H]
\centering
\includegraphics[width=0.95\textwidth]{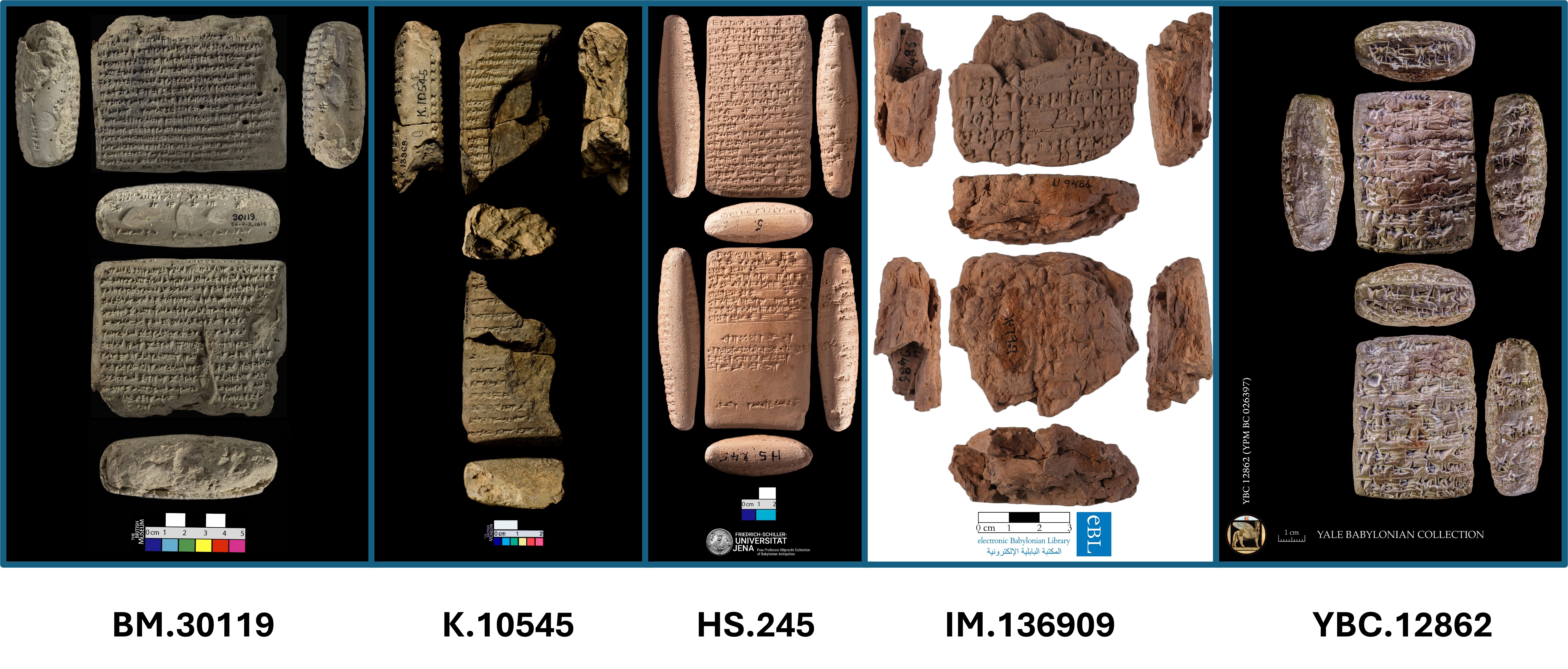}
\caption{Example images from each provenance category.}
\label{fig:provenance_examples}
\end{figure}

This dataset is aggregated as part of the Electronic Babylonian Library (eBL) platform~\citep{eBLPlatform}. Each tablet in the dataset is accompanied by transliterated sign sequences in the metadata and a corresponding annotation file that marks visible signs in the image. Metadata lists signs as ABZ-coded (\emph{Assyrisch-babylonische Zeichenliste}, the standard cuneiform sign inventory; see Section~\ref{sec:class-mapping}) or transliterated tokens in their original textual order, but does not specify visual position. The annotations provide spatial information using the format (x, y, width, height, label), where the label is the sign’s transliteration. Annotation density varies from tablet to tablet: some contain only a few identified signs while others include bounding boxes for all legible signs. Together, the metadata and annotations link textual content with its visual realisation, enabling tasks such as sign detection, recognition, and alignment between transliteration and image. 

Figure~\ref{fig:provenance_examples} shows typical tablet photographs from each provenance group, and Table~\ref{tab:provenance} summarises the museum provenance of the tablets and fragments in the corpus. The most significant contributor is the British Museum’s “Ashurbanipal Library Project”, amounting to 593 tablets photographed between 2009 and 2013. The other 422 tablets come from high-resolution photography conducted by the eBL project at the British Museum between 2018 and 2024, offering better surface appearance and fewer lighting effects. We also add 83 fragments photographed at the Iraq Museum, 336 high dynamic range (HDR) images from the Yale Babylonian Collection, and 308 images in the Hilprecht Collection located in Jena. A remaining set of 189 fragments comes from independent collections not covered by the primary categories. Altogether, the dataset comprises 1,931 photographed fragments. Individual cuneiform signs are annotated using manually defined bounding boxes, each associated with a standardised transliteration label; the bounding boxes are generated directly from the annotation data and are overlaid on the original tablet image for visualisation purposes. An example of such sign-level annotations is shown in the ground-truth panel of Figure~\ref{fig:qualitative_example_bm33535}.

Beyond museum provenance, the dataset covers the full chronological breadth of cuneiform writing traditions. Table~\ref{tab:period_timeline_quantities_percent} shows a detailed breakdown of annotated instances by historical period. Around three millennia of Mesopotamian textual culture are represented in the corpus, from early Presargonic inscriptions (0.04\% of annotations) to Late-Babylonian and early Common Era material (3.14\%). The best-represented periods are the Old Babylonian (28.90\%), Neo-Babylonian (17.61\%), Neo-Assyrian (10.66\%), Middle-Assyrian (8.99\%), Ur III (7.85\%), and Old-Assyrian periods (7.37\%). The other 13.41\% of the annotations fall within the Late Babylonian period, divided, whenever possible, into Persian, Hellenistic, and Parthian. The temporal coverage allows the dataset to capture both the diachronic variations in sign forms and the diversity of writing practices across Mesopotamian polities.

\begin{table}[h]
\centering
\caption{Provenance of tablet photographs in the dataset.}
\begin{tabular}{p{2.2cm} p{8.8cm} p{1.8cm}}
\hline
\textbf{Prefix} & \textbf{Provenance} & \textbf{Fragments} \\
\hline
K, Rm, Sm, DT & British Museum, ``Ashurbanipal Library Project'' (2009--2013) & 593 \\[1.2mm]

BM & British Museum, high-resolution in-house photography & 422 \\[1.2mm]

IM & Iraq Museum, (2018--2025) & 83 \\[1.2mm]

YBC, NBC & Yale Babylonian Collection, HDR photographs (2018--2025) & 336 \\[1.2mm]

HS & Hilprecht Collection (Jena), (2023--2025) & 308 \\[1.2mm]

Other & Other collections not covered by the categories above & 189 \\[1.2mm]

\hline
\textbf{Total} &  & \textbf{1931} \\
\hline
\end{tabular}
\label{tab:provenance}
\end{table}

\begin{table}[ht]
\centering
\caption{Chronological summary of historical periods with date ranges, annotation quantities, and their percentage contributions to our dataset.}
\label{tab:period_timeline_quantities_percent}
\begin{tabular}{l l r r}
\hline
\textbf{Period} & \textbf{Dates} & \textbf{Quantity} & \textbf{Percentage} \\
\hline
Presargonic & 2900\,BCE--2350\,BCE & 51 & 0.04\% \\
Fara & 2600\,BCE--2450\,BCE & 944 & 0.76\% \\
Sargonic & 2334\,BCE--2154\,BCE & 205 & 0.16\% \\
Ur III & 2100\,BCE--2002\,BCE & 9,778 & 7.85\% \\
Old-Babylonian & 2002\,BCE--1595\,BCE & 35,986 & 28.90\% \\
Old-Assyrian & 1950\,BCE--1850\,BCE & 9,175 & 7.37\% \\
Middle-Babylonian & 1500\,BCE--1000\,BCE & 1,384 & 1.11\% \\
Middle-Assyrian & 1363\,BCE--1050\,BCE & 11,195 & 8.99\% \\
Neo-Assyrian & 1000\,BCE--609\,BCE & 13,267 & 10.66\% \\
Neo-Babylonian & 1000\,BCE--600\,BCE & 21,926 & 17.61\% \\
Persian & 539\,BCE--331\,BCE & 4,078 & 3.28\% \\
Hellenistic & 331\,BCE--141\,BCE & 6,299 & 5.06\% \\
Parthian & 141\,BCE--100\,BCE & 6,310 & 5.07\% \\
Late-Babylonian & 600\,BCE--100\,CE & 3,906 & 3.14\% \\
\hline
{Total} & & {124,504} & {100\%} \\
\hline
\end{tabular}
\end{table}

In total, the annotation layer comprises 124,504 labelled instances. The combination of heterogeneous image sources and broad chronological coverage allows for robust evaluation of models for cuneiform character recognition, sign classification, and tablet-level reconstruction across a wide spectrum of writing traditions and material conditions.

\subsection{Properties of the Tablet Images}

The 1,931 tablet fragments in the eBL dataset exhibit substantial visual and physical diversity arising from both their state of preservation and the method used for their photography. Image resolutions vary widely, with the smallest photograph measuring \(500 \times 942\) pixels and the largest \(17{,}870 \times 11{,}409\) pixels; image heights range from \(635\) to \(25{,}029\) pixels. This broad distribution reflects differences in fragment size, photographic equipment, and digitisation protocols across collections. Two dominant background conditions are observed: \(92.2\%\) of images were captured against black backgrounds and \(7.8\%\) against white backgrounds. Beyond these measurable properties, tablet surface conditions vary considerably, including differences in clay colour, erosion, and breakage. Although not explicitly encoded in the metadata, visual inspection shows a continuum from well-preserved tablets with clearly legible inscriptions to heavily eroded fragments with edge loss, cracks, and surface abrasion. Additional variability arises from differing lighting conditions, which affect contrast, reflectivity, and shadowing. The dataset also encompasses a wide range of textual content, spanning administrative and literary traditions. Top-level genre categories include CANONICAL, ARCHIVAL, MONUMENTAL, and Other/Unavailable, with Administrative, Divination, Literature, Legal, and Technical texts being the most frequent subgenres. These categories are reported for descriptive context only, as they reflect the heterogeneity of the underlying corpus on which detection is applied. For a complete numerical summary of image dimensions, background distributions, and genre frequencies, see Tables~\ref{table: image_properties} and~\ref{tab:genre_stats}.

\begin{table}[h!]
\centering
\caption{Extreme image dimension cases and background colour statistics for the cuneiform dataset.}
\begin{tabular}{lccc}
\hline
\textbf{Category} & \textbf{ID} & \textbf{Width} & \textbf{Height} \\
\hline
Minimum Width      & K.22981   & 500    & 942    \\
Minimum Height     & K.19801   & 747    & 635    \\
Maximum Width      & YBC.4642  & 17870  & 11409  \\
Maximum Height     & BM.34035  & 10089  & 25029  \\
\hline
\multicolumn{4}{c}{\textbf{Background Colour Summary}} \\
\hline
White Backgrounds  & \multicolumn{3}{c}{151}  \\
Black Backgrounds  & \multicolumn{3}{c}{1780} \\
\hline
\end{tabular}
\label{table: image_properties}
\end{table}

\begin{table}[h!]
\centering
\caption{Distribution of top-level genres and most frequent subgenres in the cuneiform fragments.}
\begin{tabular}{l r | l r}
\hline
\textbf{Top-Level Genre} & \textbf{Count} 
& \textbf{Subgenre} & \textbf{Count} \\
\hline
CANONICAL              & 927 
& Administrative      & 305 \\
ARCHIVAL               & 742 
& Divination          & 254 \\
MONUMENTAL             & 21  
& Literature          & 221 \\
Other / Unavailable    & 241 
& Legal               & 169 \\
                       &     
& Technical           & 166 \\
                       &     
& Magic               & 108 \\
                       &     
& Celestial           & 103 \\
                       &     
& Lamentations        & 86  \\
                       &     
& Lexicography        & 67  \\
                       &     
& Hymns               & 65  \\
                       &     
& Other / Unavailable & 387 \\
\hline
\end{tabular}
\label{tab:genre_stats}
\end{table}

\subsection{Training Data and Sign Classes}
\label{subsec:training_data}

Following the class-merging strategy introduced by \citet{Cobanoglu2024}, we train two DETR models: a 173-class model, which is the label set after merging infrequent categories, and a 106-class model aligned with the baseline.
For the 173‑class setting, all sign categories with fewer than 90 annotated instances are merged into a single \texttt{UnclearSign} class as in the previous setting; for the 106‑class setting, only the 106 most frequent categories are retained, matching the class scheme of the baseline.
The \texttt{UnclearSign} category therefore includes both signs that were explicitly annotated as unclear or illegible by experts, and otherwise identifiable sign classes whose instance counts fall below the frequency threshold.

\begin{figure}[H]
\centering
\includegraphics[width=0.95\textwidth]{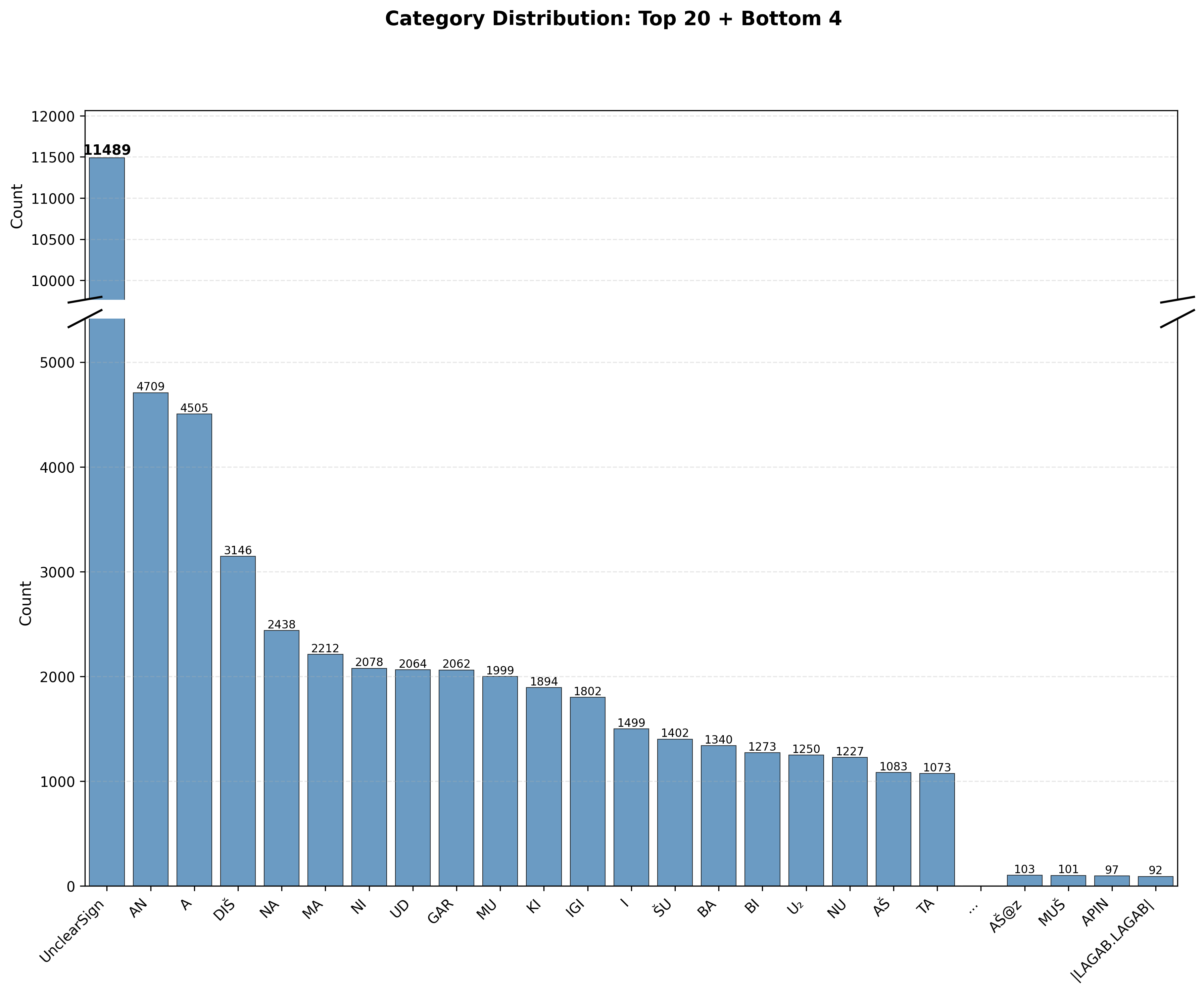}
\caption{Histogram of sign frequencies in the training set. All classes with fewer than 90 instances are merged into the \texttt{UnclearSign} category.}
\label{fig:class_histogram}
\end{figure}

\paragraph{Class distribution and imbalance.}
The distribution of sign frequencies in the training data is highly imbalanced. As illustrated in Figure~\ref{fig:class_histogram}, a small number of sign classes account for a large proportion of all annotated instances, while the majority of classes occur far less frequently. The most prominent example is the \texttt{UnclearSign} category, which aggregates both inherently ambiguous signs and low-frequency sign classes, and consequently forms the largest class in the dataset. Beyond this dominant category, only a limited set of signs (e.g., \texttt{AN}, \texttt{A}, \texttt{DIŠ}, \texttt{NA}) reach several thousand instances, whereas many classes appear close to the lower frequency bound.

This long-tailed distribution reflects a structural property of cuneiform corpora, where common administrative and grammatical signs are reused extensively, while many lexical or context-specific signs remain rare. Such imbalance poses challenges for supervised training and evaluation, and is consistent with observations reported in earlier studies on cuneiform sign detection and OCR \citep{Cobanoglu2024}.

\paragraph{Motivation for reduced-class scheme.}
We consider two class granularities for complementary reasons. The 173-class configuration preserves broader sign coverage after merging only categories with fewer than 90 annotations into \texttt{UnclearSign}, which is useful for downstream analyses that require finer philological distinctions. The 106-class configuration follows the label scheme of \citet{Cobanoglu2024}, providing a direct baseline comparison while also reducing the effective label space and thereby easing the class-imbalance problem. Quantitative comparisons between the two settings are reported in Section~\ref{sec:result}.

\paragraph{Data split and comments.}
 Following the baseline setup, only tablet fragments for which annotated signs account for more than 60\% of the total visible signs on the tablet were included in the training. This threshold is computed by comparing the number of manually annotated bounding boxes against the count of reading and logogram signs in the transliteration, excluding sections marked as broken away. Specifically, for each fragment, we extract all non-broken reading and logogram tokens from the transliteration metadata via the eBL database, and retain the fragment only if the ratio of annotations to transliteration-derived sign count exceeds 0.6. This ensures that training data exhibit sufficient annotation coverage relative to the expected textual content.

\begin{figure}[H]
\centering
\includegraphics[width=0.70\textwidth]{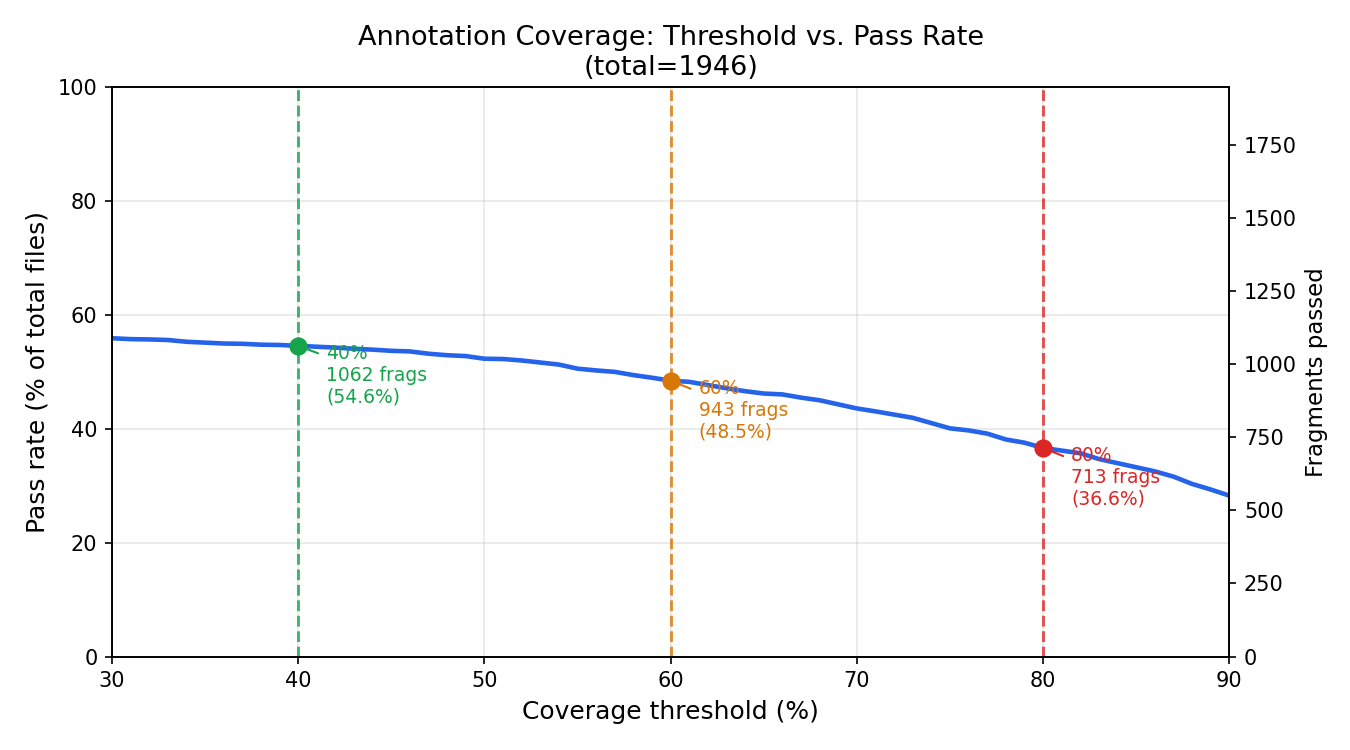}
\caption{Retained fragments as a function of the annotation coverage threshold (computed on a snapshot of 1,946 fragments). The decline is gradual below 60\% but steepens markedly above it.}
\label{fig:coverage_curve}
\end{figure}

The choice of 60\% is motivated empirically by Figure~\ref{fig:coverage_curve}, computed on a near-identical snapshot of 1,946 fragments. The pass rate declines gradually between 40\% and 60\% (1,062 to 943 fragments) but drops sharply above 60\% (713 fragments at 80\%), indicating a natural inflection point that balances annotation quality against dataset size.

Applying the 60\% threshold to the full dataset of 1,949 fragments yields 947 retained fragments. Subsequent tablet-side extraction (Section~\ref{sec:side-extraction}) expands this to 1,552 individual side images, which are split into 1,452 training images and 100 test images. Two splitting strategies were employed: (i) a random split, and (ii) a stratified split that samples proportionally from each period–provenance combination to ensure distributional similarity with the baseline evaluation setup.

\subsubsection*{Data preprocessing}
\begin{itemize}
    \item \textbf{UnclearSign grouping:} All sign labels with fewer than 90 instances are merged into the \texttt{UnclearSign} category to reduce extreme class imbalance.
    \item \textbf{Image resizing:} The model uses multi-scale random resizing (480--800 × 1,333) during training for data augmentation and fixed resizing to 1,333 × 800 during inference, both preserving aspect ratio.
    \item \textbf{Tablet-side extraction:} Multi-surface tablet photographs are automatically segmented into individual tablet sides prior to inference, see Section~\ref{sec:side-extraction}.
\end{itemize}

\subsection{Class Mapping}
\label{sec:class-mapping}

Sign labels in the dataset are standardised using the ABZ numbering system, following the conventions of the Electronic Babylonian Library (eBL). ABZ numbers are taken from Borger’s \emph{Assyrisch-babylonische Zeichenliste} \citep{Borger1988ABZ}, which provides a widely accepted canonical inventory of cuneiform signs and their variants in the first millennium BCE. Using ABZ identifiers allows consistent handling of sign labels and easy mapping between different periods, corpora, and editorial traditions. In practice, annotated sign names or transliteration tokens are mapped to their corresponding ABZ entries. As described in Section~\ref{subsec:training_data}, sign classes with insufficient training instances (fewer than 90 examples) are merged into the \texttt{UnclearSign} category, along with signs that were explicitly annotated as illegible or ambiguous during the manual annotation process.

\section{Method}

\subsection{Model: DETR}

We adopt the Deformable DETR architecture with a ResNet-50 backbone and Hungarian matching for end-to-end object detection. Training follows the configuration in \verb|configs/detr.py| with modifications only to the number of classes and size of the dataset, to allow a direct comparison with the baseline results in \citet{Cobanoglu2024}.
The essential hyperparameters used in all experiments are summarised in Table~\ref{tab:reproducibility}.

\begin{table}[ht]
\centering
\caption{Essential hyperparameters used across experiments.}
\label{tab:reproducibility}
\begin{tabular}{ll}
\hline
\textbf{Parameter} & \textbf{Setting} \\
\hline
Model & Deformable DETR (two-stage, box refinement) \\
Backbone & ResNet-50 (ImageNet pretrained) \\
Number of Queries & 300 \\
Dataset format & COCO annotation format \\
Batch Size & 2 \\
Optimiser & AdamW \\
Learning Rate & $2 \times 10^{-4}$ \\
Weight Decay & $1 \times 10^{-4}$ \\
Learning Rate Schedule & Multi-step decay (epoch 40, $\gamma = 0.1$) \\
Training Epochs & 1000 \\
Loss Functions & Focal (classification), L1 (box), generalised IoU (GIoU) \\
Assignment Method & Hungarian matching \\
Data Augmentation & Multi-scale resize, random crop, flip, colour and blur augmentations \\
\hline
\end{tabular}
\end{table}

The classification head in Deformable DETR uses focal loss by default; all loss-function parameters were left at their defaults and no architectural changes were made beyond adjusting the number of output classes. Focal loss is well-suited to the long-tailed class distribution in our data: as noted by \citet{Cobanoglu2024}, focal loss can counter the extreme class imbalance present in cuneiform sign datasets.

The choice of DETR comes from its advantages over two-stage methods: it does not require large cropped-sign training data, and its end-to-end formulation makes it faster at inference.

All models were trained on a single NVIDIA V100 GPU with 16 GB of memory. The batch size was set to 2 to fit within GPU memory constraints, and a typical training run required approximately 2 days. The training framework also supports multi-GPU execution, allowing the batch size to be increased to reduce training time. Training data were organised in the Common Objects in Context (COCO) dataset format \citep{Lin2014COCO}, and model performance was evaluated using standard COCO metrics.

\subsection{Inference Pipeline Extensions}

\subsubsection{Automatic Tablet Side Extraction}
\label{sec:side-extraction}

Unlike the fully annotated examples used for training, many eBL images contain multiple tablet surfaces or partial tablets in a single photograph. Since training assumed one tablet side per image, we introduced an automatic segmentation step to extract individual tablet sides from photographs.

First we apply adaptive thresholding to detect non-background contours against the near-uniform black or white photographic backdrop and use their enclosing rectangles as candidate bounding boxes. We then discard implausible candidates that are extremely small (less than 0.005\% of the full image area), extremely large (more than 90\%), or extremely thin (aspect ratio greater than 100), merge boxes that lie within 5 pixels of one another nearby, and retain at most the 12 largest merged regions. Each retained region is finally cropped with 10 pixels of padding on all sides. These threshold values were chosen empirically on the basis of conventions in the museum photography layout.

This preprocessing ensures that the inference inputs resemble the scale and framing of the training data, reducing prediction errors.

\begin{figure}[H]
\centering
\includegraphics[width=0.45\textwidth]{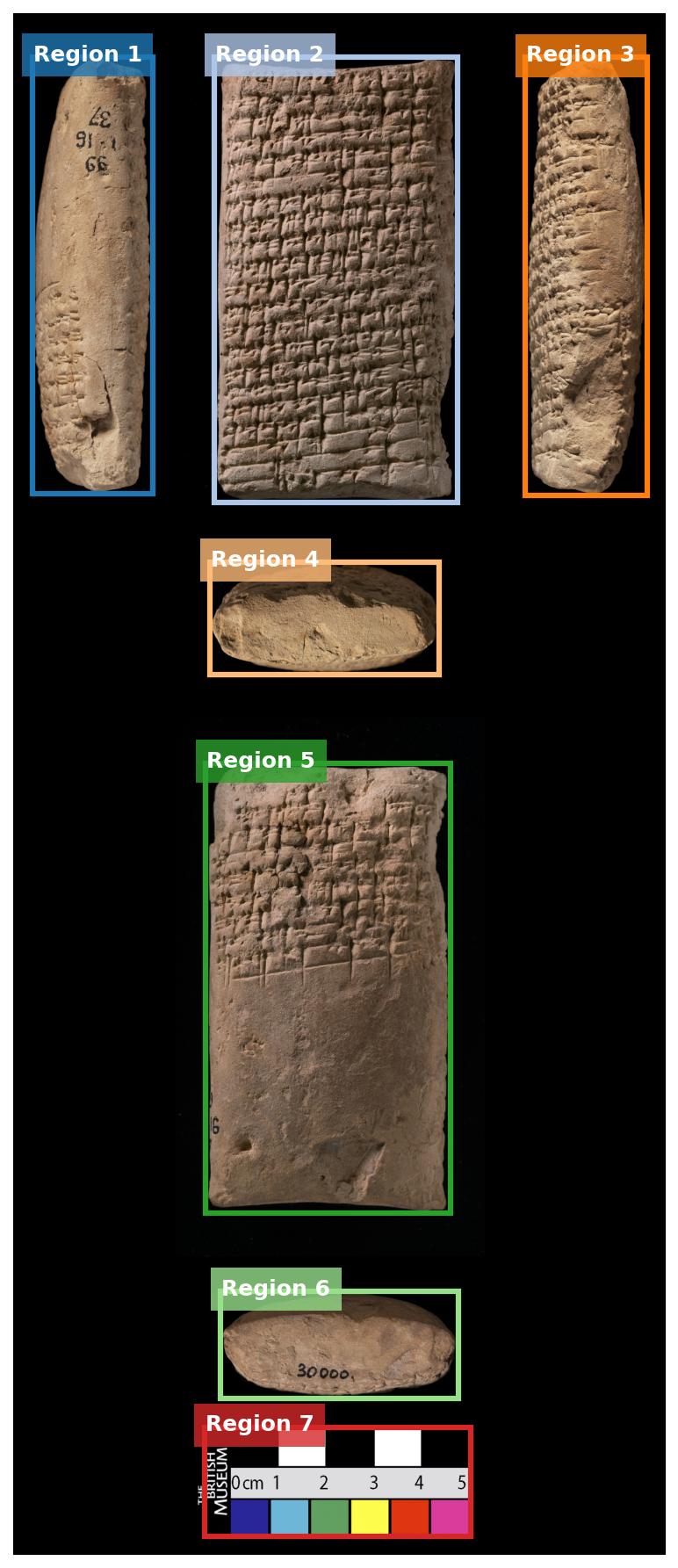}
\caption{Automatic tablet side extraction example. Multiple tablet sides are identified within a single image and marked as separate regions (Region 1--7). Each region is subsequently cropped and processed independently for sign detection.}
\label{fig:tablet_side_extraction_overview}
\end{figure}

Figure~\ref{fig:tablet_side_extraction_overview} illustrates the tablet side extraction process. The original image may contain multiple tablet surfaces or fragments. Detected tablet regions are indicated by bounding boxes and are subsequently cropped and processed individually and then passed to the DETR model for sign detection.

\begin{figure}[H]
\centering
\includegraphics[width=0.95\textwidth]{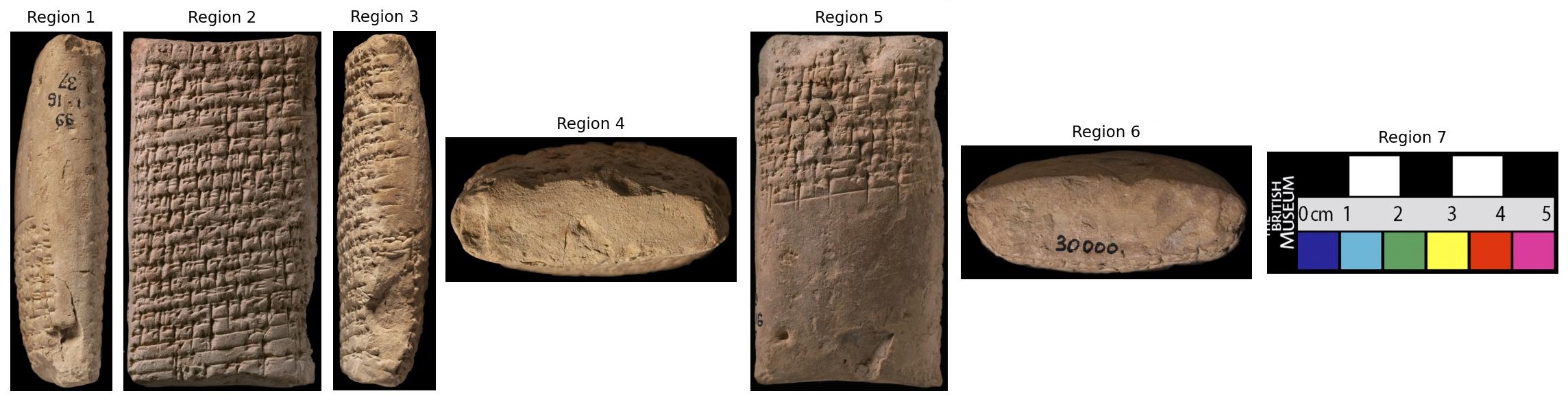}
\caption{Cropped tablet sides obtained from the segmentation step. Each region includes a 10-pixel padding around the detected tablet boundary and is used as input for sign detection. Non-tablet elements such as colour bars (e.g., Region~7) may also be extracted but do not yield sign detections and have no impact on the final output.}
\label{fig:tablet_side_extraction_crops}
\end{figure}

\paragraph{Validation.}
To assess segmentation quality, we conducted a manual validation on a random sample of photographs drawn directly from the image database.
Automatic validation against existing ground-truth annotations was considered but ultimately not used as the primary measure: the dataset's tablet-side annotations are both incomplete (many photographs have only a subset of visible sides annotated) and of inconsistent quality (box boundaries vary substantially across annotators), making intersection-over-union (IoU)-based metrics an unreliable indicator of segmentation correctness.

Manual validation was carried out by visually inspecting the segmentation output for each sampled fragment.
Results on 100 sampled fragments are shown in Table~\ref{tab:side_extraction_validation}.
The recall of 98.6\% indicates that nearly all tablet sides are successfully detected; the seven missed sides are consistently small or partially obscured fragments.
Note that all false detections correspond to scale rulers or image labels (the case of Region~7 in Figure~\ref{fig:tablet_side_extraction_crops}). These regions produce no sign detections and do not affect the final output, though they consume a small amount of inference compute.

\begin{table}[h]
\centering
\caption{Manual validation of tablet-side extraction, with 95\% Wilson confidence
intervals reported for the binomial proportions.}
\label{tab:side_extraction_validation}
\begin{tabular}{lc}
\hline
\textbf{Metric} & \textbf{Value} \\
\hline
Predicted sides & 586 \\
TP (correctly detected)      & 518 \\
FN (missed)     & 7 \\
FP (false det.) & 68 \\
Recall          & 98.6\% [97.2\%,\,99.3\%] \\
Precision       & 88.4\% [85.5\%,\,90.7\%] \\
\hline
\end{tabular}
\end{table}

\subsubsection{Line Grouping}
\label{sec:line-grouping}

To group detected signs into horizontal text lines, we employ DBSCAN clustering with a custom anisotropic distance metric that strongly de-emphasises horizontal separation and clusters primarily by vertical position. For any two detected signs $A$ and $B$ with centres $(x_A, y_A)$ and $(x_B, y_B)$, the distance is defined as:

\begin{equation}
\label{eq:dbscan-distance}
\mathrm{Dist}(A, B) = \sqrt{\lambda\,(x_A - x_B)^2 + (y_A - y_B)^2},
\end{equation}

where $\lambda \ll 1$ is a weighting parameter that suppresses the contribution of horizontal displacement. Before clustering, both coordinates are normalised by the average sign size (mean of average sign width and height across the image), making the distance metric scale-invariant and robust to variation in tablet resolution and sign dimensions across the dataset.

DBSCAN \citep{ester1996dbscan} is then applied to the normalised, $\lambda$-scaled coordinates. In this formulation, $\varepsilon$ defines the radius threshold in normalised sign-size units: two signs are considered neighbours if their scaled distance falls below $\varepsilon$. The $\texttt{min\_samples}$ parameter specifies the minimum number of points required within the $\varepsilon$-neighbourhood for a point to be classified as a core point; signs with fewer neighbours are labelled as noise. Setting $\texttt{min\_samples} = 2$ means that a sign must have at least one other sign within its $\varepsilon$-neighbourhood to belong to a cluster, which effectively suppresses truly isolated spurious detections while still allowing small line fragments to form valid clusters. Each resulting cluster corresponds to a detected text line. Within each cluster, signs are sorted left to right by their horizontal centre coordinate, and clusters are ordered top to bottom by their mean vertical position.

Compared to a fixed pixel-threshold approach, this formulation offers two key advantages. First, normalisation by average sign size allows a single set of parameters to generalise across images of substantially different scales and resolutions. Second, the anisotropic distance function accommodates moderate horizontal overlap between adjacent lines—common in dense cuneiform layouts—while remaining robust to noise detections that are vertically isolated from the main text body. The three parameters ($\varepsilon$, $\texttt{min\_samples}$, $\lambda$) are tuned jointly with the detection confidence threshold via n-gram matching on a held-out validation set, as described in Section~\ref{sec:n-gram}.

\begin{figure}[H]
\centering
\includegraphics[width=0.45\textwidth]{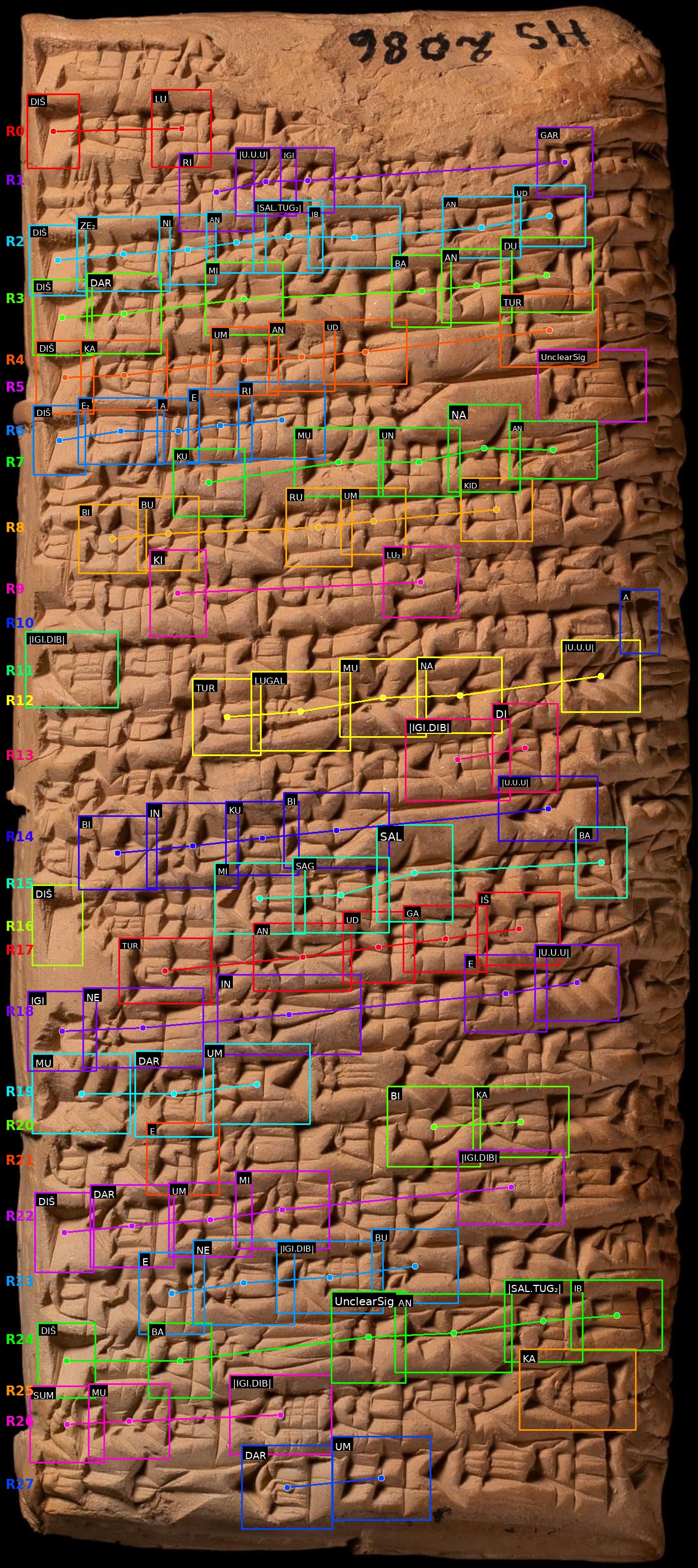}
\caption{Example of DBSCAN-based line grouping on tablet HS.9807. Each detected text line is assigned a distinct colour; signs belonging to the same line are connected and labelled with their row identifier on the left. Signs within each row are sorted left to right by horizontal position.}
\label{fig:line_grouping_example}
\end{figure}

Figure~\ref{fig:line_grouping_example} shows an example of the line grouping results on a cuneiform tablet.

This produces a structured sequence representation of the detected signs that approximates the line-based layout of cuneiform writing, which is essential for the n-gram–based textual evaluation described in Section~\ref{sec:n-gram}.

\subsection{Evaluation with n-gram Matcher}
\label{sec:n-gram}
We employ an n-gram matching approach for two critical purposes in this work: first, to determine optimal detection thresholds through parameter tuning on large-scale sign inference on the eBL dataset; second, to conduct a comprehensive evaluation of our final model against all annotated sign sequences available in the database.

Textual similarity is measured using the union of 1-, 2-, and 3-gram overlaps between model predictions and manually curated sign sequences from transliterations. Let $A$ and $B$ denote the sets of 1-, 2-, and 3-grams extracted from the predicted and reference sign sequences, respectively. An n-gram matcher \citep{Simonjetz2024} computes:

\begin{equation}
\label{eq:overlap}
\text{Overlap}(A,B) = |A \cap B|,
\end{equation}

and

\begin{equation}
\label{eq:match-score}
\text{Score}(A,B) = \frac{|A \cap B|}{\min(|A|, |B|)}.
\end{equation}

In this framework, the overlap metric reflects the absolute degree of n-gram correspondence without normalisation by sequence length, whereas the match score measures the proportion of shared n-grams relative to the shorter of the two sequences. Equation~(\ref{eq:overlap}) defines the overlap, and equation~(\ref{eq:match-score}) the match score; the match score is the primary indicator, and the overlap is used as a complementary measure.

This approach is robust to incomplete or partially ordered detection results and follows the method used in the eBL platform for fragment matching. High match scores indicate strong textual correspondence between predicted and reference sign sequences. Ground truth annotations are obtained directly from the eBL database via their public application programming interface (API).

\section{Results}
\label{sec:result}

\subsection{Accuracy performance}
\label{sec:performance}

Detection performance is evaluated using COCO-style metrics. Average Precision (AP) is reported over IoU thresholds ranging from 0.50 to 0.95, together with fixed-threshold scores AP$_{50}$ and AP$_{75}$. Average Recall (AR) is computed under the same IoU range at maximum-detection cutoffs of 100, 300, and 1000 detections per image, following the evaluation configuration used by \citet{Cobanoglu2024}. Table~\ref{tab:combined_metrics} reports our three evaluation settings together with the 106-class old-test-set baseline from \citep{Cobanoglu2024}.

On the newly constructed test dataset, the 106-class configuration achieves slightly higher aggregate performance than the 173-class model across most metrics, particularly for AP$_{50}$ and AR. This improvement is expected, not caused by the smaller number of classes in itself, but because the reduced-class formulation merges a larger number of infrequent sign categories into the \textit{UnclearSign} class, thereby reducing classification ambiguity and increasing the effective sample size of that category. As \textit{UnclearSign} is already the most frequent label in the dataset, this consolidation makes many previously difficult distinctions easier to classify and leads to gains in aggregate detection scores.

\begin{figure}[H]
\centering
\includegraphics[width=0.75\textwidth]{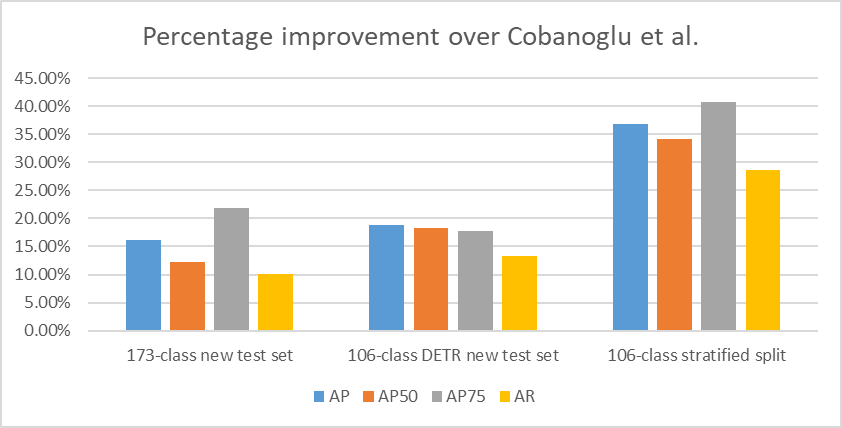}
\caption{Percentage improvement (\%) of the proposed 173-class, 106-class on new test set and 106-class on stratified split test set (fair comparison) DETR models over the baseline method of \citet{Cobanoglu2024} across AP, AP$_{50}$, AP$_{75}$, and AR metrics.}
\label{fig:percentage_improvement}
\end{figure}

To better isolate the contribution of model improvements from data distribution effects, we also evaluate a 106-class model trained using a stratified sampling strategy. Rather than randomly splitting the dataset, we construct the test set by sampling proportionally from each period–provenance combination observed in the original test data. This ensures that the training and test distributions more closely match those used in prior work, enabling a fairer assessment of methodological advances. Under this stratified configuration, the 106-class model achieves an AP of 0.312 and AP$_{50}$ of 0.515, while the corresponding 106-class old-test-set baseline reported by \citet{Cobanoglu2024} achieves AP = 0.228 and AP$_{50}$ = 0.384. Because the original and newly constructed test sets differ in composition, we use this stratified setting as the basis for the fair-comparison improvements summarised in Figure~\ref{fig:percentage_improvement}.

Across all evaluated settings, our DETR variants outperform the baseline of \citet{Cobanoglu2024} on the reported COCO-style metrics. Under the matched 106-class comparison based on the stratified split, the improvement ranges from 28\% to 37\%, as summarised in Figure~\ref{fig:percentage_improvement}.

Despite these gains, the 173-class model offers substantially richer categorical coverage by explicitly modelling 67 additional sign types. While this increased granularity comes at the cost of slightly lower overall metrics, it is essential for downstream applications that rely on fine-grained sign distinctions, such as philological analysis and corpus-level textual reconstruction. For this reason, the 173-class model is used for large-scale inference over the full Electronic Babylonian Library (eBL) tablet collection.

\begin{table*}[t]
\centering
\caption{COCO-style detection evaluation metrics for different class numbers and test datasets. All metrics follow the standard COCO evaluation protocol. Metrics for small objects are omitted as no valid ground-truth instances are present in the corresponding scale range of the new test split. The stratified split maintains the same 100-image test set size while ensuring proportional representation of period–provenance combinations. The final column reproduces the 106-class results reported by \citet{Cobanoglu2024} on their original test split.}
\label{tab:combined_metrics}
\begin{tabular}{lcccc}
\hline
\textbf{Metric} &
\textbf{173 classes} &
\textbf{106 classes} &
\textbf{106 classes} &
\textbf{106 classes} \\
&
\textbf{(new test set)} &
\textbf{(new test set)} &
\textbf{(stratified split)} &
\textbf{(old test set)} \\
\hline
\multicolumn{5}{l}{\textit{Average Precision (AP)}} \\
AP            & 0.265 & 0.271 & 0.312 & 0.228 \\
AP$_{50}$     & 0.431 & 0.454 & 0.515 & 0.384 \\
AP$_{75}$     & 0.302 & 0.292 & 0.349 & 0.248 \\
\hline
\multicolumn{5}{l}{\textit{AP across scales}} \\
AP$_{small}$  & --    & --    & 0.050 & 0.138 \\
AP$_{medium}$ & 0.131 & 0.128 & 0.217 & 0.213 \\
AP$_{large}$  & 0.282 & 0.292 & 0.321 & 0.238 \\
\hline
\multicolumn{5}{l}{\textit{Average Recall (AR)}} \\
AR$_{max=100}$  & 0.316 & 0.325 & 0.369 & 0.287 \\
AR$_{max=300}$  & 0.316 & 0.325 & 0.369 & 0.287 \\
AR$_{max=1000}$ & 0.316 & 0.325 & 0.369 & 0.287 \\
\hline
\multicolumn{5}{l}{\textit{AR across scales}} \\
AR$_{small}$  & --    & --    & 0.050 & 0.237 \\
AR$_{medium}$ & 0.140 & 0.140 & 0.250 & 0.268 \\
AR$_{large}$  & 0.336 & 0.349 & 0.389 & 0.289 \\
\hline
\end{tabular}
\end{table*}

\clearpage

\begin{figure}[H]
\centering
\includegraphics[width=0.90\textwidth]{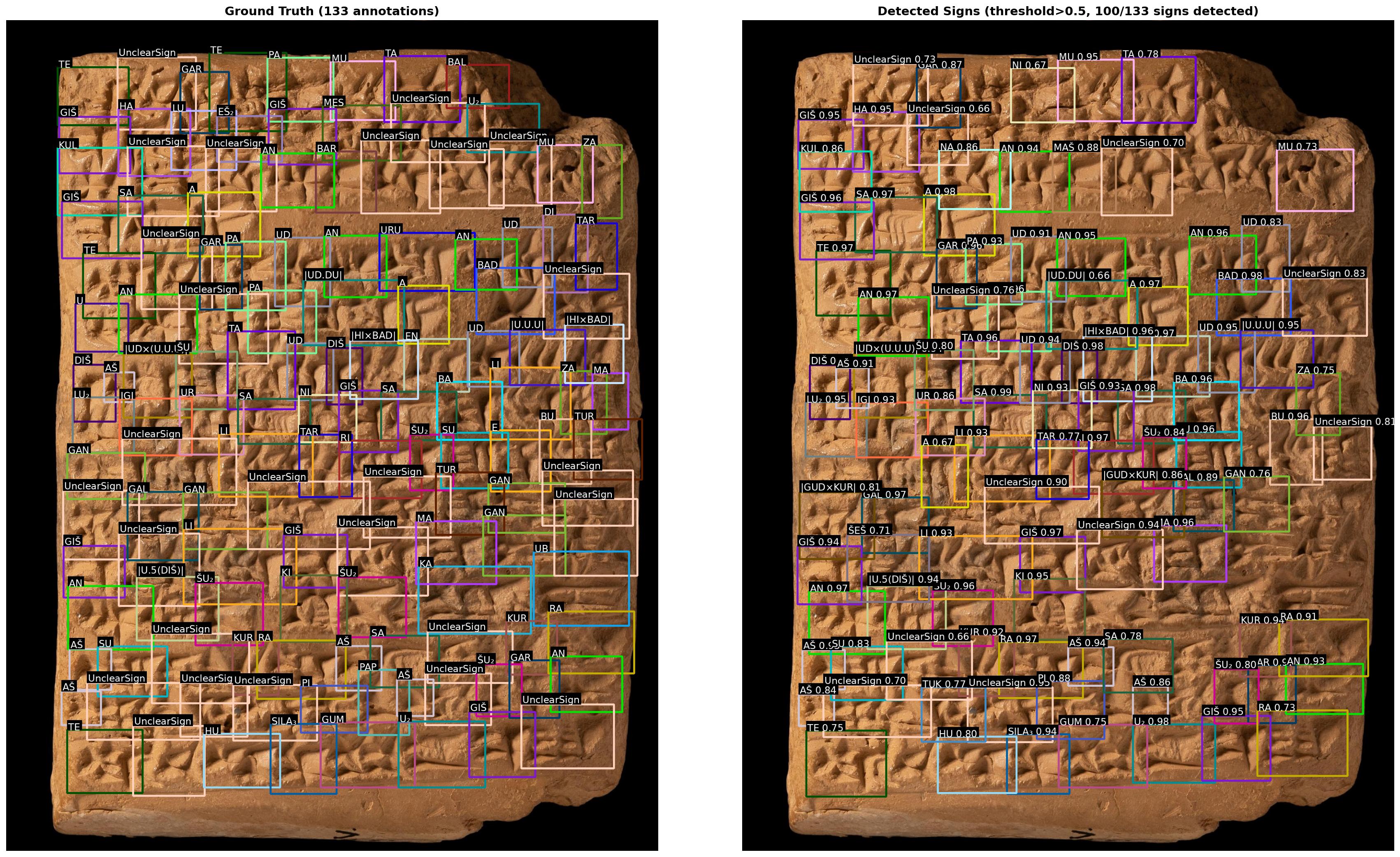}
\caption{Qualitative comparison of sign detection results for tablet BM.33535. Left: ground truth annotations. Right: DETR predictions with detection threshold 0.5; this tablet achieves AP$_{50}$ = 0.69.}
\label{fig:qualitative_example_bm33535}
\end{figure}

Figure~\ref{fig:qualitative_example_bm33535} presents a qualitative comparison between ground truth annotations (left) and model predictions (right) for tablet BM.33535, using a detection confidence threshold of 0.5. The model achieves an AP$_{50}$ of 0.69 on this tablet, demonstrating accurate localisation and classification of well-preserved signs, while remaining robust to moderate layout variation.

\begin{figure}[H]
\centering
\includegraphics[width=0.90\textwidth]{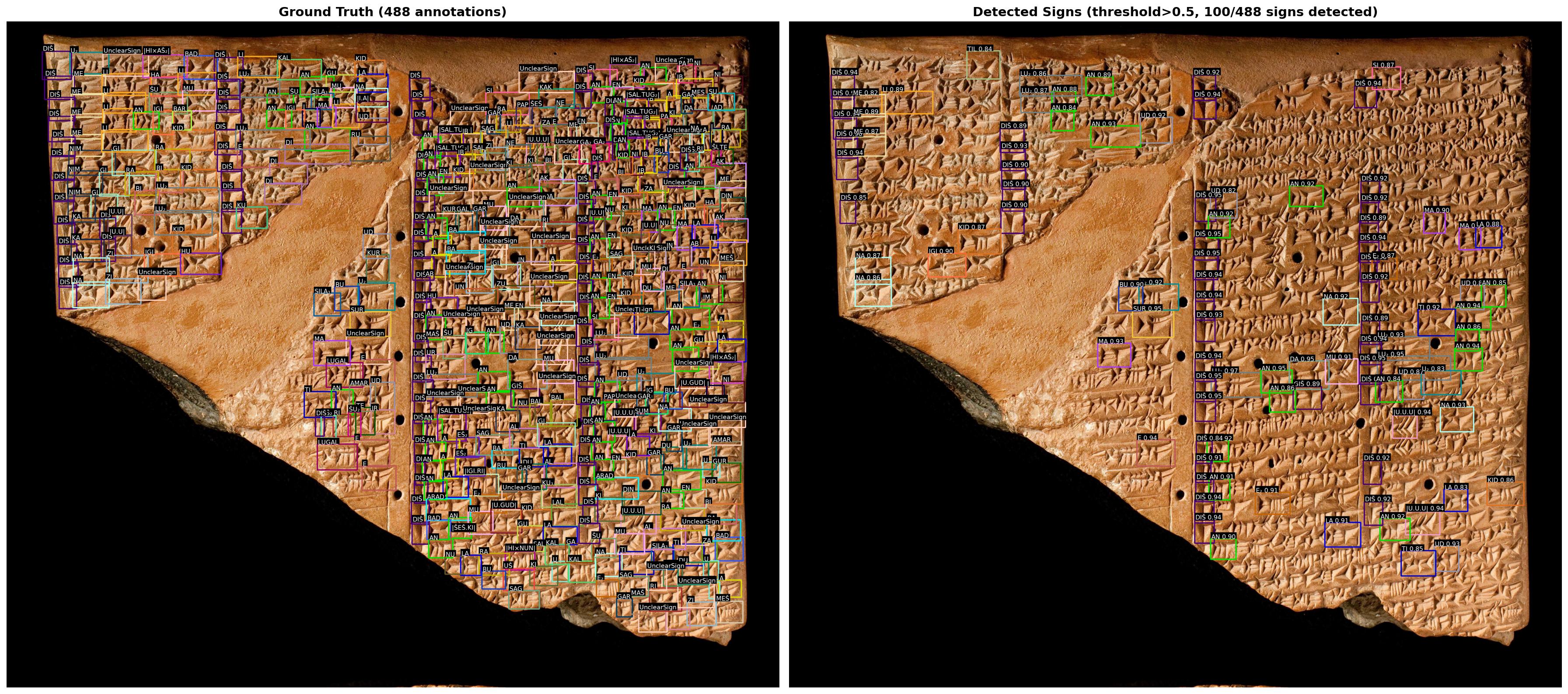}
\caption{Failure-case qualitative comparison for tablet K.4426. Left: ground truth annotations (488 annotated signs). Right: DETR predictions with detection threshold 0.5, recovering only 100 signs.}
\label{fig:qualitative_failure_k4426}
\end{figure}

Figure~\ref{fig:qualitative_failure_k4426} shows a complementary failure case on tablet K.4426. At the same detection threshold of 0.5, the model recovers only 100 of 488 annotated signs, illustrating the detector's reduced recall on large tablets with dense sign layouts.

\noindent To examine how detection quality varies with training frequency, Table~\ref{tab:per_class_ap} reports AP@0.5:0.95 for the ten most and ten least frequent sign classes in the 173-class training set (173-class model, new test set). High-frequency classes achieve consistent AP values in the range 0.26–0.38, with the exception of \textit{UnclearSign} (AP = 0.122), whose low score reflects the inherent ambiguity of this catch-all category rather than a detector failure. Among the least frequent classes (92–104 training instances), most still achieve reasonable AP values (0.23–0.30); only a small number of classes with fewer than 100 training instances—\textit{MUŠ} (0.090), \textit{IL} (0.076), and \textit{APIN} (0.000)—show markedly degraded performance. Overall, the results indicate that the long-tailed class distribution does not cause a significant accuracy problem across the majority of sign types; the outlier cases are confined to the very bottom of the frequency distribution.

\begin{table}[t]
\centering
\caption{Per-class AP@0.5:0.95 for the ten most frequent (left) and ten least frequent (right) sign classes in the 173-class training set. \textit{UnclearSign} is a catch-all pseudo-class grouping ambiguous signs. $\dagger$~Class absent from the test split, AP not computable.}
\label{tab:per_class_ap}
\begin{tabular}{lrr@{\hspace{2em}}lrr}
\hline
\multicolumn{3}{c}{\textbf{Top-10 by Training Frequency}} &
\multicolumn{3}{c}{\textbf{Bottom-10 by Training Frequency}} \\
\textbf{Class} & \textbf{Train} & \textbf{AP} &
\textbf{Class} & \textbf{Train} & \textbf{AP} \\
\hline
\textit{UnclearSign} & 10{,}740 & 0.122 & \textit{|LAGAB.LAGAB|}$^\dagger$ &  92 & -- \\
AN                   &  4{,}376 & 0.265 & MUŠ                             &  93 & 0.090 \\
A                    &  4{,}240 & 0.348 & APIN                            &  96 & 0.000 \\
DIŠ                  &  2{,}922 & 0.300 & IL                              &  98 & 0.076 \\
NA                   &  2{,}297 & 0.337 & 6(DIŠ)                          & 100 & 0.254 \\
MA                   &  2{,}068 & 0.325 & AŠ@z                            & 100 & 0.135 \\
NI                   &  1{,}973 & 0.287 & GI$_4$                          & 101 & 0.232 \\
UD                   &  1{,}955 & 0.289 & GAR$_3$                         & 103 & 0.269 \\
GAR                  &  1{,}952 & 0.284 & MUŠ$_3$                         & 103 & 0.300 \\
MU                   &  1{,}870 & 0.377 & |IGI.RI|                        & 104 & 0.230 \\
\hline
\end{tabular}
\end{table}

\subsection{Large-scale Inference on the eBL Tablet Collection}
\label{sec:large_scale_inference}

Using the trained 173-class DETR model, we performed large-scale inference on the full tablet image collection available in the Electronic Babylonian Library (eBL) database~\citep{eBLPlatform}. This step aims to produce a consistent, automatically generated inventory of detected signs across the corpus, suitable for downstream quantitative and textual analyses.

\paragraph{Inference procedure.}
All tablet images were first processed using the automatic tablet-side extraction pipeline described in Section~\ref{sec:side-extraction}, ensuring that each inference input corresponds to a single tablet surface. The extracted tablet sides were then passed through the 173-class DETR model to obtain sign-level bounding boxes and classification scores.

For each detected instance, the model outputs a sign label and a confidence score. To transform raw detections into structured sign sequences, an additional set of parameters was applied:
\begin{itemize}
    \item a \emph{detection confidence threshold} controlling which predicted signs are retained;
    \item three \emph{line grouping parameters} ($\varepsilon$, $\texttt{min\_samples}$, $\lambda$) governing the DBSCAN-based row detection described in Section~\ref{sec:line-grouping}.
\end{itemize}

\paragraph{Threshold selection via n-gram matching.}
Rather than selecting these thresholds heuristically, we determine suitable values empirically using the n-gram matching framework introduced earlier. A subset of tablets in which more than 40\% of the text was manually annotated was used as the validation set. For each tablet in this subset, predicted sign sequences were compared against curated transliteration-derived sign sequences using the n-gram overlap and match score metrics.

We evaluated combinations of detection-confidence thresholds and line-grouping parameters, analysing the trade-off between match score and overlap. Lower detection thresholds yield higher overlap by retaining more signs, but at the cost of increased false positives and reduced textual coherence. Conversely, higher thresholds suppress too many true detections, leading to a reduction in overlap. Under the constraint that overlap should not fall below an acceptable level, we selected a detection confidence threshold of $0.8$, with line grouping parameters $\varepsilon = 0.35$, $\texttt{min\_samples} = 2$, and $\lambda = 0.006$. The full parameter sweep results are provided in the Appendix.

This configuration yielded the most balanced performance in terms of textual overlap and match-score stability across the validation tablets. The resulting detections were subsequently converted into ABZ-formatted sign labels and integrated into the database for further analysis.

\begin{figure}[H]
\centering
\includegraphics[width=0.90\textwidth]{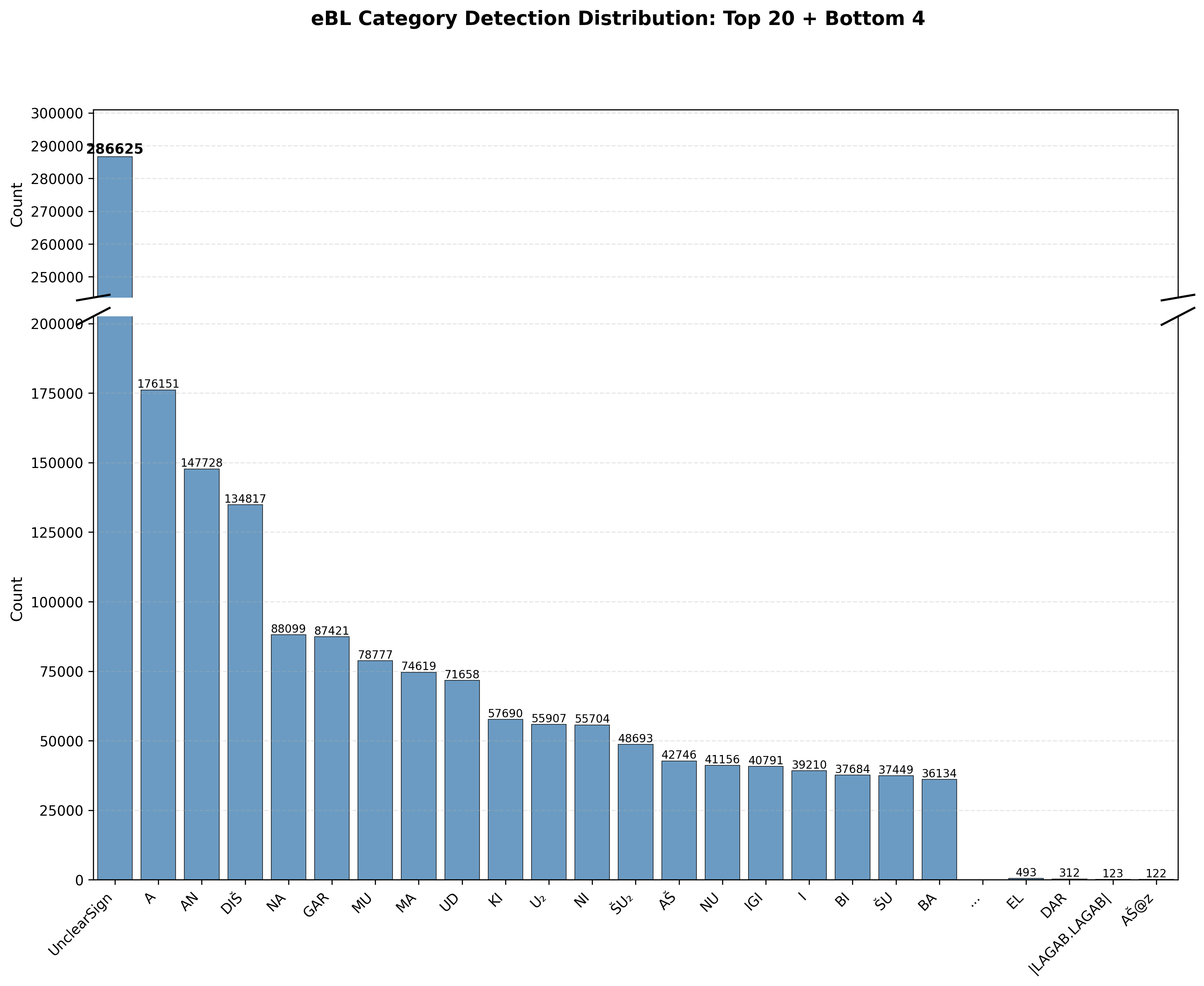}
\caption{Frequency distribution of detected signs across the full eBL tablet collection using the 173-class DETR model.}
\label{fig:sign_frequency_full}
\end{figure}

Applying the selected inference configuration to the full eBL tablet collection produces a large-scale distribution of detected sign types. Figure~\ref{fig:sign_frequency_full} summarises the frequency of each sign class after inference using the 173-class model.

Inference was carried out on {87,668 tablet fragments}, yielding a total of {2,892,350 detected sign instances}. As illustrated in Figure~\ref{fig:sign_frequency_full}, the resulting sign frequency distribution is broadly consistent with that of the training data (Figure~\ref{fig:class_histogram}), exhibiting a similar long-tailed structure dominated by common administrative and grammatical signs. This qualitative agreement suggests that the model preserves global sign usage patterns when applied at scale, while minor deviations are expected due to corpus composition and preservation variability.

\subsection{N-gram Match Result}
\label{sec:ngram-match-result}

To quantitatively assess the quality of the large-scale inference, we evaluated the alignment between the automatically detected signs and the manual transliterations available in the eBL database.
To ensure unbiased evaluation, we excluded the 381 fragments used in the training set from this analysis.
Since not all remaining fragments in the corpus possess a complete or robust ground-truth transliteration suitable for n-gram comparison, this evaluation was performed on a subset of 26{,}790 fragments with sufficient textual metadata to compute meaningful overlap and match scores.

\begin{figure}[H]
\centering
\includegraphics[width=0.95\textwidth]{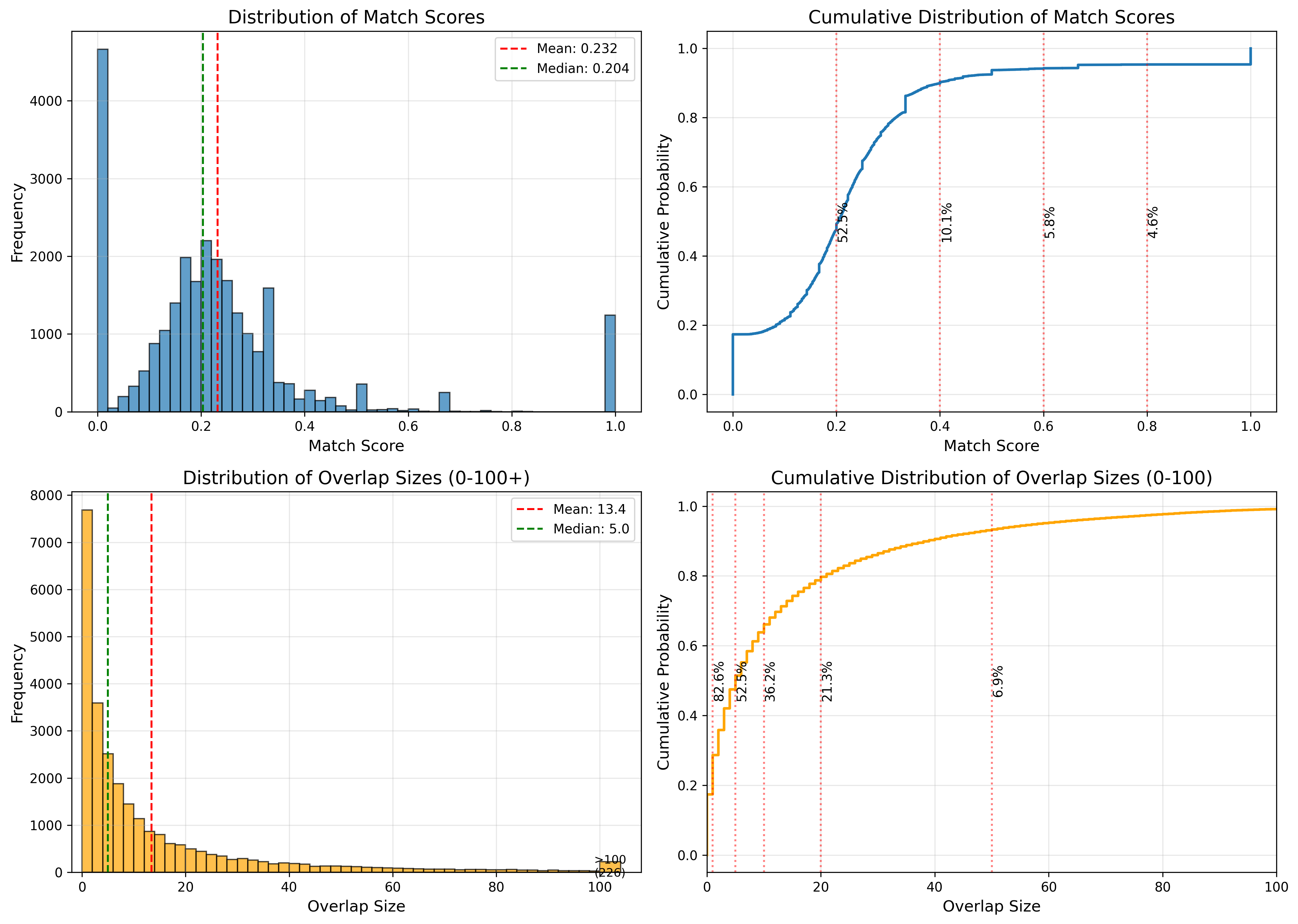}
\caption{Statistical distribution of N-gram matching results on 26{,}790 fragments (excluding the training set) at detection threshold $0.8$, $\varepsilon = 0.35$, $\texttt{min\_samples} = 2$, and $\lambda = 0.006$. Top row: histogram and cumulative distribution of Match Scores (Mean: 0.232, 95\% CI [0.230, 0.234]; Median: 0.204, 95\% CI [0.202, 0.206]). Bottom row: histogram and cumulative distribution of Overlap Sizes (Mean: 13.4, 95\% CI [13.0, 13.7]; Median: 5.0, 95\% CI [5.0, 5.0]). Confidence intervals are 95\% bootstrap intervals based on 10{,}000 resamples.}
\label{fig:ngram_distributions}
\end{figure}

Here, Match Score denotes the normalised overlap between the predicted and reference 1-, 2-, and 3-gram sets, whereas Overlap Size counts the absolute number of shared n-grams. Figure~\ref{fig:ngram_distributions} presents the full distribution and cumulative density of both metrics for this evaluation, and Table~\ref{tab:ngram_anchors} places the headline result in context using two lower anchors and one upper anchor.

\begin{table}[H]
\centering
\caption{Interpretive anchors for the n-gram evaluation. For UB we report n-gram recall rather than raw Match Score, because the raw score saturates near 1.0 after collapsing out-of-vocabulary signs into \texttt{UnclearSign}.}
\label{tab:ngram_anchors}
\begin{tabular}{p{3.0cm}p{5.1cm}cc}
\hline
\textbf{Anchor} & \textbf{Primary metric} & \textbf{Mean} & \textbf{95\% CI} \\
\hline
Headline model & Match Score (predictions vs. own transliterations) & 0.232 & [0.230,\,0.234] \\
LB1 random-label baseline & Match Score (labels resampled from empirical detection frequencies) & 0.070 & [0.068,\,0.070] \\
LB2 cross-fragment shuffle & Match Score (predictions matched to another fragment's transliteration) & 0.110 & [0.109,\,0.113] \\
UB 173-class vocabulary ceiling & Recall of reference n-grams after label-space collapse & 0.910 & [0.908,\,0.912] \\
\hline
\end{tabular}
\end{table}

The gap between LB2 and the headline result (0.122) reflects fragment-specific recognition signal beyond generic corpus-level co-occurrence statistics, while UB shows that roughly 9\% of reference n-grams are excluded by the current 173-class label space alone.

The quantitative analysis reveals the following insights.

\paragraph{Match Score Distribution.}
The distribution of Match Scores (Figure~\ref{fig:ngram_distributions}, top row) has a mean of $0.232$ (95\% CI [0.230, 0.234]) and a median of $0.204$ (95\% CI [0.202, 0.206]).
While the central mass of the distribution indicates a consistent alignment capability, we observe distinct outliers at the extremes ($0.0$ and $1.0$), with approximately 10\% of the data clustering at these boundaries.
The spike at a perfect score of $1.0$ is largely attributable to the trade-off between score and overlap size: fragments with very short overlapping sequences (low overlap) are statistically more likely to achieve a perfect match, as the probability of a localised error decreases with sequence length.
Conversely, the accumulation of scores at $0.0$ reflects fragments where detection failed significantly.
These failures are typically caused by illegible handwriting, unique epigraphic patterns not represented in the training distribution, extreme scale variations, or adverse photographic conditions (e.g., poor lighting or low contrast).
The tightness of the bootstrap intervals around both the mean and the median (width well below $0.005$) reflects the large evaluation sample and indicates that the reported central tendencies are stable summaries of the underlying distribution.

\paragraph{Overlap Size Distribution.}
The Overlap Size distribution (Figure~\ref{fig:ngram_distributions}, bottom row) is heavily right-skewed, with a mean intersection of $13.4$ signs (95\% CI [13.0, 13.7]) and a median of $5.0$ signs. This distribution accurately reflects the fragmentary nature of the underlying corpus, in which most objects are small, broken pieces containing limited text.
However, a notable tail exists with well-preserved tablets on which the model successfully recovered substantial portions of the text.
The narrow CI around the mean confirms that the gap between the mean ($13.4$) and the median ($5.0$) is a genuine property of the corpus rather than a consequence of estimation noise.

\subsection{Ablation Studies}
\label{sec:ablation}

To assess the contribution of individual components in the proposed pipeline, we conduct two ablation studies.

\subsubsection{With vs. without tablet-side extraction}
\label{sec:ablation_tablet_side}

To evaluate the contribution of the tablet-side extraction step to overall detection quality, we train an additional 173-class DETR model on the same dataset but \emph{without} applying tablet-side extraction: the full, unprocessed tablet photographs are used directly as both training and inference inputs. Both conditions are evaluated on the same 100-image test set, drawn from the same data pool as the training data (i.e., the new test set described in Section~\ref{sec:performance}); the only difference between conditions is the preprocessing applied to the training and test images. Table~\ref{tab:ablation_tablet_side} compares COCO detection metrics for the two conditions. Removing tablet-side extraction leads to a substantial drop across all metrics: AP falls from 0.265 to 0.107, and AP$_{50}$ from 0.431 to 0.207. This confirms that preprocessing the image to isolate individual tablet surfaces is a critical step: it removes background clutter, normalises the spatial context for each surface, and ensures that the detector focuses on the relevant portion of the image rather than wasting capacity on non-sign regions such as scale bars and mounting fixtures.

\begin{table}[h]
\centering
\caption{Ablation of tablet-side extraction. Both models use 173 sign classes
and are evaluated on the same 100-image new test set, with a reported two-sample $z$-test on AP.}
\label{tab:ablation_tablet_side}
\begin{tabular}{lcc}
\hline
\textbf{Metric} & \textbf{With extraction} & \textbf{Without extraction} \\
\hline
\multicolumn{3}{l}{\textit{Average Precision (AP)}} \\
AP            & 0.265 & 0.107 \\
AP$_{50}$     & 0.431 & 0.207 \\
AP$_{75}$     & 0.302 & 0.096 \\
\hline
\multicolumn{3}{l}{\textit{AP across scales}} \\
AP$_{small}$  & --    & --    \\
AP$_{medium}$ & 0.131 & 0.054 \\
AP$_{large}$  & 0.282 & 0.117 \\
\hline
\multicolumn{3}{l}{\textit{Average Recall (AR)}} \\
AR$_{max=1000}$ & 0.316 & 0.146 \\
\hline
\multicolumn{3}{l}{\textit{AR across scales}} \\
AR$_{small}$  & --    & --    \\
AR$_{medium}$ & 0.140 & 0.068 \\
AR$_{large}$  & 0.336 & 0.156 \\
\hline
\multicolumn{3}{l}{\textit{Significance on AP}} \\
$\Delta$AP, $z$, $p$ & \multicolumn{2}{c}{$+0.158$,\quad $z=4.47$,\quad $p < 10^{-5}$} \\
\hline
\end{tabular}
\end{table}

\subsubsection{Fixed-threshold vs. DBSCAN line grouping}
\label{sec:ablation_line_grouping}

To demonstrate the improvement brought by the proposed line grouping method, we re-computed the n-gram match scores using the original fixed pixel-threshold approach, in which lines are separated by a single uniform y-axis pixel threshold applied across the entire tablet.
This approach has two tunable parameters: the detection-confidence threshold and the y-axis pixel-separation threshold.
With all other conditions held constant, the optimal configuration of this baseline was found to be detection threshold $=0.7$ and y-threshold $=0.35$, yielding a mean match score of $0.205$ and a median of $0.182$.

Compared with the proposed DBSCAN-based line grouping (Mean: 0.232, Median: 0.204; see Section~\ref{sec:ngram-match-result}), the new method achieves a clear improvement, primarily by eliminating the effects of scale variation and line skew, which in turn improves the accuracy of row assignment.
The full distributions of match scores and overlap sizes under the fixed-threshold baseline are reported in the Appendix.

\subsection{Limitations}
\label{sec:limitations}

Despite the encouraging results, several limitations of the proposed inference pipeline should be acknowledged.

\paragraph{Absence of semantic and linguistic context.}
The existing detection system operates purely at the visual and structural level. Even though the identified signs are clustered into rough textual lines, the detector treats each sign prediction independently, without sequence modelling. So there are no semantic, syntactic or lexical constraints incorporated during inference. In natural cuneiform texts, sign co-occurrence is highly structured: certain signs or sign sequences appear together with far higher probability than would be expected from independent frequency statistics alone.

Although adding linguistic context might enhance the accuracy of detection and the coherence of the sequence, it would also add powerful priors that would cause the output to follow familiar or well-established patterns. Assyriologically, it can be beneficial to maintain a relatively free process of detection, as it allows the possibility of recognising unusual combinations of signs, rare formulae or textual patterns that were previously underrepresented. The question of how semantic direction and exploratory openness can be balanced is also a key area of future research.

\paragraph{Tablet damage and preservation variability.}

A substantial proportion of the analysed tablets exhibit varying degrees of physical damage. This includes, but is not limited to: (i) fragments preserving only a small portion of the original tablet, (ii) severely irregular or broken edges, and (iii) surfaces affected by partial erosion, cracks, or abrasions. In such cases, sign shapes may be incomplete or distorted.

The model does not reliably classify these damaged signs into the \texttt{UnclearSign} category, often producing confident but incorrect predictions. This behaviour both reduces inference accuracy and introduces noise into the effective training signal, as visually ambiguous regions are not explicitly distinguished from well-preserved signs.

Large tablets with dense sign layouts also remain difficult. Figure~\ref{fig:qualitative_failure_k4426} illustrates this failure mode on K.4426, where only 100 of 488 annotated signs are recovered at threshold 0.5.

\paragraph{Line grouping robustness.}
Although the DBSCAN-based line grouping is more robust than a simple fixed-threshold approach, it still has limitations. The clustering parameters are globally tuned and do not adapt to local variations within a single tablet. Tablets with highly irregular line spacing, curved writing surfaces, or multi-column layouts may not be fully captured by the current formulation. More adaptive or learning-based layout modelling could improve sequence reconstruction in such cases, but was beyond the scope of the present work.

Overall, these limitations reflect deliberate design choices favouring simplicity, scalability, and interpretability. Addressing them will likely require tighter integration of visual, linguistic, and layout-aware modelling approaches.

\section{Conclusions}

We extend a DETR-based pipeline for cuneiform sign detection. Beyond reproducing the core components, we introduce several inference-time enhancements that make the model's output more interpretable for textual tasks, including tablet-side segmentation and line grouping. The proposed system improves COCO-style detection performance by 28--37\% over the matched 106-class baseline of \citet{Cobanoglu2024}. At corpus scale, the 173-class model was applied to 87{,}668 tablet fragments and produced 2,892,350 sign detections. The n-gram similarity evaluation further connects visual detection to textual structure, offering a bridge toward full transliteration workflows.

As the Electronic Babylonian Library continues to expand in both scale and annotation coverage, the availability of large, systematically detected sign inventories opens new directions for computational Assyriology. Future work may integrate sequence-level modelling and multimodal Transformer-based architectures that jointly reason over visual sign detections and textual context. These developments would enable more robust contextual analysis and large-scale fragment search. They would also support closer integration between visual evidence and philological interpretation.

\clearpage

\section*{Appendix}
\label{appendix}

\subsection{N-gram Parameter Sweep}
\label{app:param-sweep}

To select the detection confidence threshold, DBSCAN row-grouping parameters ($\varepsilon$, $\texttt{min\_samples}$), and line-height scale $\lambda$, we conducted a random search over the joint parameter space evaluated against a validation set of fragments with partial manual transliterations.
For each sampled configuration, predicted sign sequences were compared to ground-truth transliterations using an n-gram match score and overlap size as joint criteria.
Configurations were ranked by average match score; to ensure robustness, high-scoring but isolated combinations, those whose metric values deviated substantially from neighbouring configurations in parameter space, were excluded in favour of stable regions where both score and overlap vary smoothly.

The overlap drops sharply once the threshold exceeds $0.8$, indicating that higher thresholds suppress too many true detections.
The threshold was therefore set to $0.8$ to maintain an acceptable overlap while still filtering low-confidence predictions.
The remaining parameters were chosen from the stable high-score region. The final configuration is: detection confidence threshold $0.8$, $\varepsilon = 0.35$, $\texttt{min\_samples} = 2$, and $\lambda = 0.006$.

\begin{figure}[H]
\centering
\includegraphics[width=0.95\textwidth]{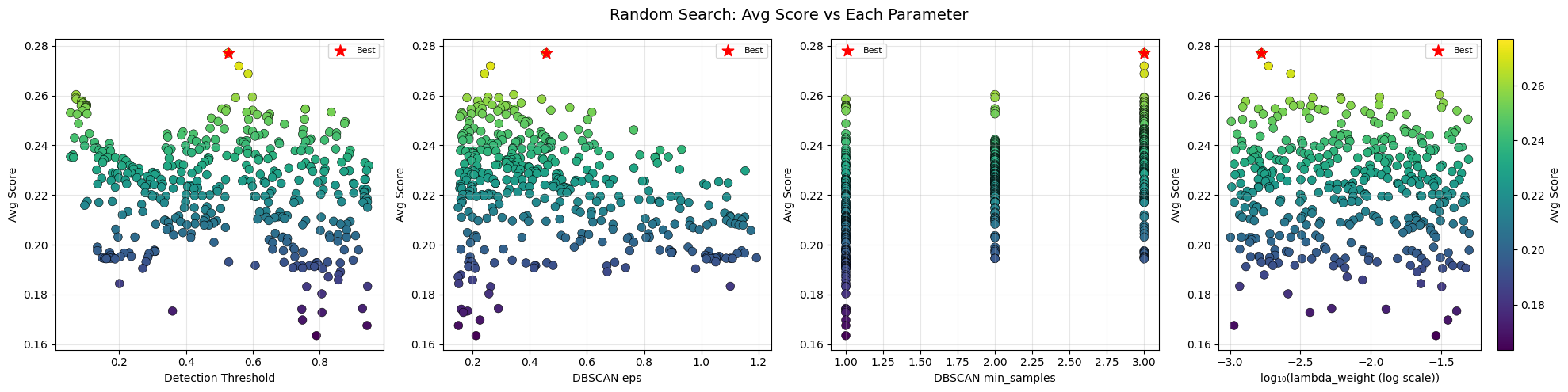}
\caption{Average n-gram match score versus each of the four parameters across all random search trials.}
\label{fig:param-sweep-1}
\end{figure}

\begin{figure}[H]
\centering
\includegraphics[width=0.95\textwidth]{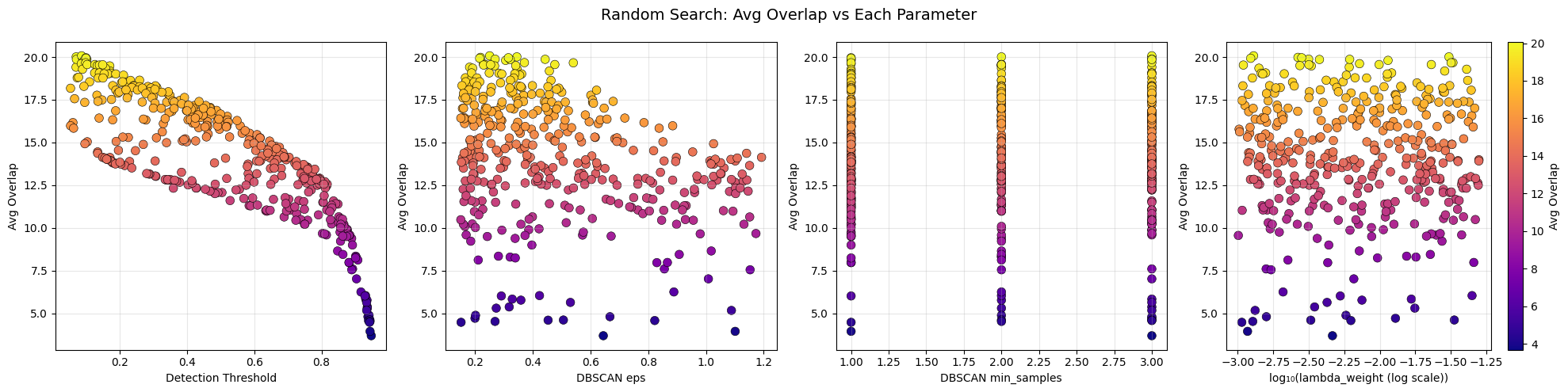}
\caption{Average n-gram overlap size versus each parameter.}
\label{fig:param-sweep-2}
\end{figure}

\begin{figure}[H]
\centering
\includegraphics[width=0.95\textwidth]{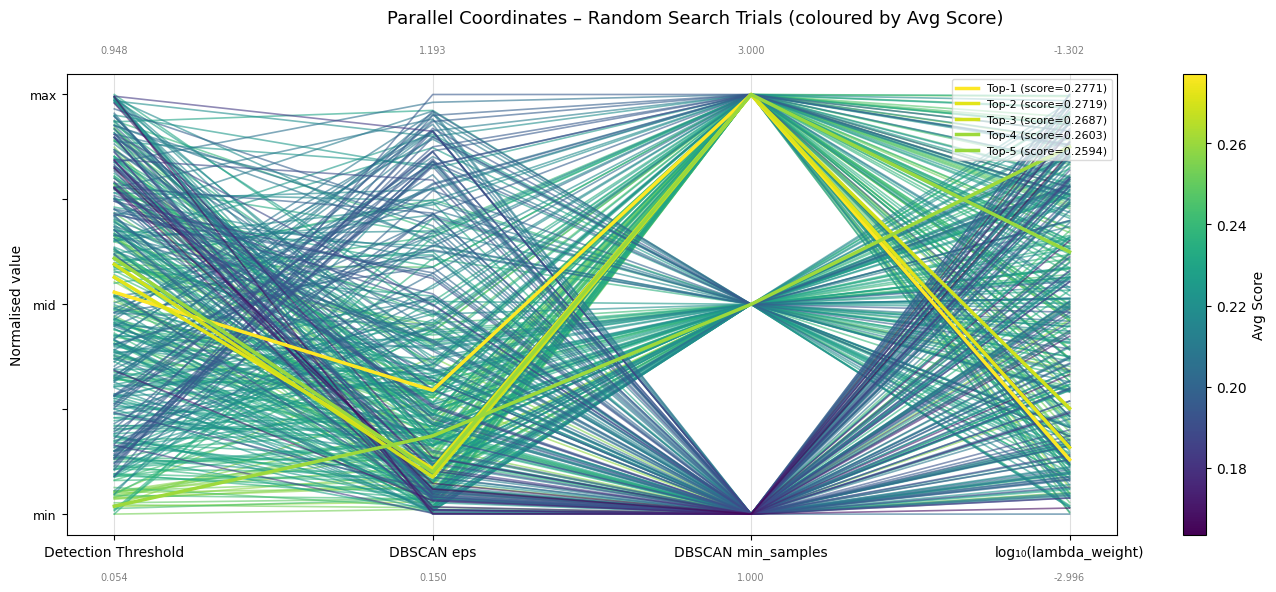}
\caption{Parallel coordinates of all random search trials coloured by average match score.}
\label{fig:param-sweep-3}
\end{figure}

\subsection{Fixed-threshold Line Grouping Results}
\label{app:fixed-threshold-results}

This section reports the n-gram matching distributions obtained with the fixed pixel-threshold line-grouping baseline, evaluated on the same 26{,}790 fragments used for the main n-gram evaluation in the main article.
The figure mirrors the layout of the main n-gram distribution figure, enabling direct comparison between the two line-grouping strategies under otherwise identical inference and evaluation conditions.

\begin{figure}[H]
\centering
\includegraphics[width=0.90\textwidth]{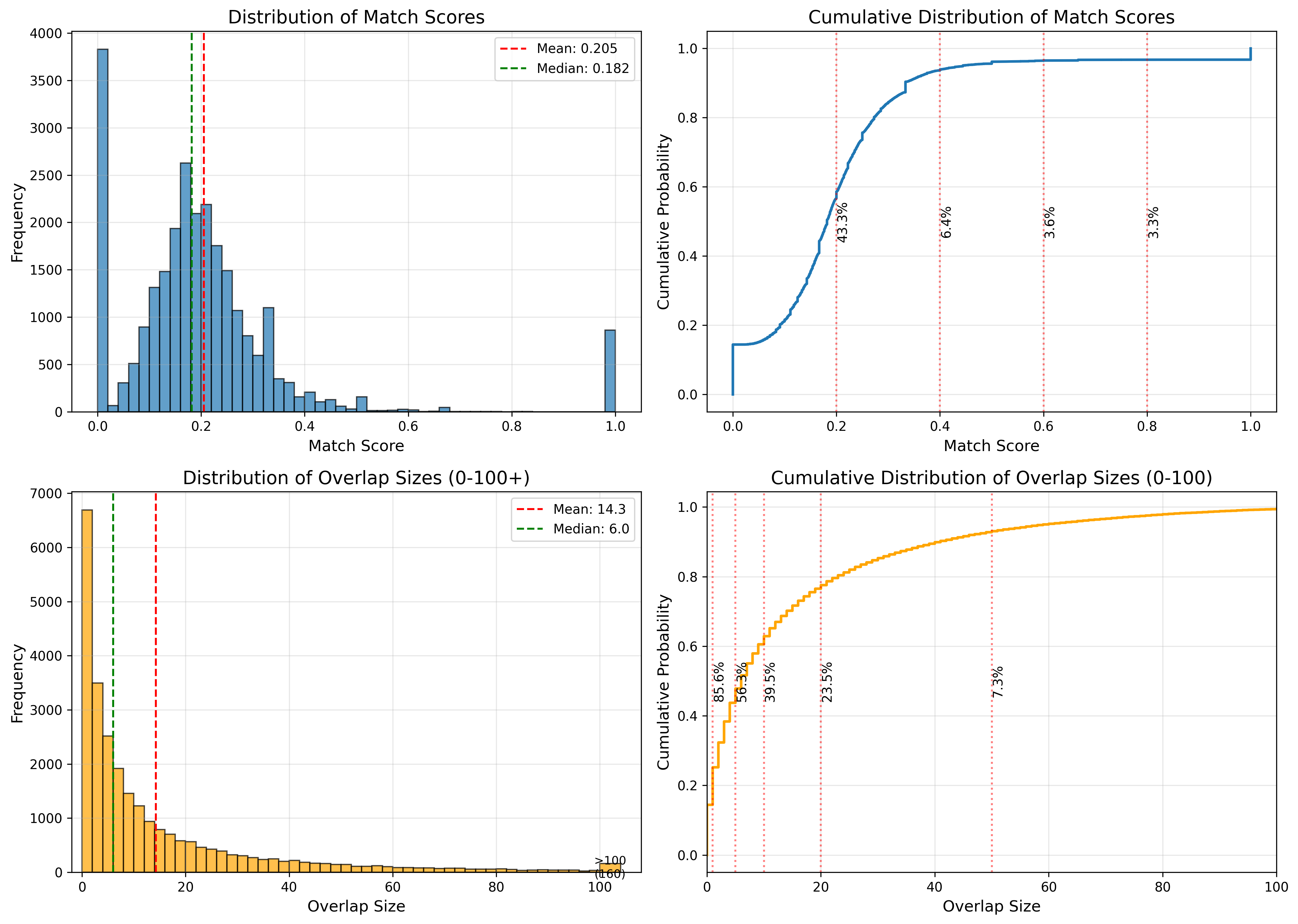}
\caption{Statistical distribution of N-gram matching results on 26,790 fragments (excluding training set) using the fixed pixel-threshold line grouping baseline at detection threshold $0.7$ and y-threshold $0.35$.
Top row: Histogram and cumulative distribution of Match Scores (Mean: 0.205, Median: 0.182).
Bottom row: Histogram and cumulative distribution of Overlap Sizes (Mean: 14.3, Median: 6.0).
}
\label{fig:ngram_distributions_fixed}
\end{figure}

\theendnotes

\section*{Figure legends}
\label{sec:figure-legends}

\noindent\textbf{Figure~\ref{fig:provenance_examples}.} Example images from each provenance category.\par\medskip

\noindent\textbf{Figure~\ref{fig:class_histogram}.} Histogram of sign frequencies in the training set. All classes with fewer than 90 instances are merged into the \texttt{UnclearSign} category.\par\medskip

\noindent\textbf{Figure~\ref{fig:coverage_curve}.} Retained fragments as a function of the annotation coverage threshold (computed on a snapshot of 1,946 fragments). The decline is gradual below 60\% but steepens markedly above it.\par\medskip

\noindent\textbf{Figure~\ref{fig:tablet_side_extraction_overview}.} Automatic tablet side extraction example. Multiple tablet sides are identified within a single image and marked as separate regions (Region 1--7). Each region is subsequently cropped and processed independently for sign detection.\par\medskip

\noindent\textbf{Figure~\ref{fig:tablet_side_extraction_crops}.} Cropped tablet sides obtained from the segmentation step. Each region includes a 10-pixel padding around the detected tablet boundary and is used as input for sign detection. Non-tablet elements such as colour bars (e.g., Region~7) may also be extracted but do not yield sign detections and have no impact on the final output.\par\medskip

\noindent\textbf{Figure~\ref{fig:line_grouping_example}.} Example of DBSCAN-based line grouping on tablet HS.9807. Each detected text line is assigned a distinct colour; signs belonging to the same line are connected and labelled with their row identifier on the left. Signs within each row are sorted left to right by horizontal position.\par\medskip

\noindent\textbf{Figure~\ref{fig:percentage_improvement}.} Percentage improvement (\%) of the proposed 173-class, 106-class on new test set and 106-class on stratified split test set (fair comparison) DETR models over the baseline method of \citet{Cobanoglu2024} across AP, AP$_{50}$, AP$_{75}$, and AR metrics.\par\medskip

\noindent\textbf{Figure~\ref{fig:qualitative_example_bm33535}.} Qualitative comparison of sign detection results for tablet BM.33535. Left: ground truth annotations. Right: DETR predictions with detection threshold 0.5; this tablet achieves AP$_{50}$ = 0.69.\par\medskip

\noindent\textbf{Figure~\ref{fig:qualitative_failure_k4426}.} Failure-case qualitative comparison for tablet K.4426. Left: ground truth annotations (488 annotated signs). Right: DETR predictions with detection threshold 0.5, recovering only 100 signs.\par\medskip

\noindent\textbf{Figure~\ref{fig:sign_frequency_full}.} Frequency distribution of detected signs across the full eBL tablet collection using the 173-class DETR model.\par\medskip

\noindent\textbf{Figure~\ref{fig:ngram_distributions}.} Statistical distribution of N-gram matching results on 26{,}790 fragments (excluding the training set) at detection threshold $0.8$, $\varepsilon = 0.35$, $\texttt{min\_samples} = 2$, and $\lambda = 0.006$. Top row: histogram and cumulative distribution of Match Scores (Mean: 0.232, 95\% CI [0.230, 0.234]; Median: 0.204, 95\% CI [0.202, 0.206]). Bottom row: histogram and cumulative distribution of Overlap Sizes (Mean: 13.4, 95\% CI [13.0, 13.7]; Median: 5.0, 95\% CI [5.0, 5.0]). Confidence intervals are 95\% bootstrap intervals based on 10{,}000 resamples.\par\medskip

\section*{Data availability}
\label{sec:data-availability}

To support reproducibility, trained models and derived detection results are made publicly available.
The trained weights for both the 173-class and 106-class DETR models are released via Zenodo
(DOI: \url{https://zenodo.org/records/17395154}).
In addition, OCR detection results generated by the 173-class model on the full eBL tablet image collection are provided at \url{https://zenodo.org/records/18863050}.

The complete project codebase, including model training, inference, and evaluation pipelines,
is publicly accessible as an open-source repository at \url{https://github.com/ElectronicBabylonianLiterature/cuneiform-ocr}.
Code for dataset construction and data preprocessing, covering annotation filtering, class mapping,
and tablet-side extraction, is available at \url{https://github.com/ElectronicBabylonianLiterature/cuneiform-ocr-data}.

The n-gram matcher used for textual similarity evaluation is available at
\url{https://github.com/ElectronicBabylonianLiterature/ngram-matcher}.

The annotated cuneiform sign dataset analysed in this study is publicly available; see \citet{lewenstein2026large} for the dataset description and the eBL platform (\citealp{eBLPlatform}) for the underlying tablet metadata and transliterations.

\section*{Competing interests}
\label{sec:competing-interests}
The author(s) declare no competing interests.

\section*{Ethical approval}
\label{sec:ethical-approval}
This article does not contain any studies with human participants performed by any of the authors.

\section*{Informed consent}
\label{sec:informed-consent}
This article does not contain any studies with human participants performed by any of the authors.

\section*{Author contributions}
\label{sec:author-contributions}

Wentao Che conceived and led the project, developed the object detection training and inference pipeline, conducted the primary experiments, and wrote the manuscript. Esteban Garcés Arias designed and performed the statistical analyses and uncertainty quantification, expanded the methodological positioning and related-work analysis, and contributed substantially to manuscript revision. Asim Niaz contributed to manuscript preparation, methodological feedback, figures, tables, and visualizations. Andreas Bender provided machine-learning guidance, supervision, and manuscript feedback. Enrique Jiménez initiated and supervised the project, provided domain expertise, and coordinated the submission. All authors contributed to the discussion, revision, and approval of the final manuscript.


\begin{thebibliography}{}

\bibitem[Ahmed, 2012]{ahmed2012online}
Ahmed, K.~K. (2012).
\newblock Online sumarians cuneiform detection based on symbol structural
  vector algorithm.
\newblock {\em Journal of the College of Education for Women}, 23(2).

\bibitem[Bogacz and Mara, 2022]{bogacz2022digital}
Bogacz, B. and Mara, H. (2022).
\newblock Digital assyriology---advances in visual cuneiform analysis.
\newblock {\em Journal on Computing and Cultural Heritage (JOCCH)},
  15(2):1--22.

\bibitem[Borger, 1988]{Borger1988ABZ}
Borger, R. (1988).
\newblock {\em Assyrisch-babylonische Zeichenliste}.
\newblock Neukirchen-Vluyn, 4 edition.

\bibitem[Carion et~al., 2020]{carion2020endtoendobjectdetectiontransformers}
Carion, N., Massa, F., Synnaeve, G., Usunier, N., Kirillov, A., and Zagoruyko,
  S. (2020).
\newblock End-to-end object detection with transformers.

\bibitem[Chammas et~al., 2018]{chammas2018handwriting}
Chammas, E., Mokbel, C., and Likforman-Sulem, L. (2018).
\newblock Handwriting recognition of historical documents with few labelled
  data.
\newblock In {\em 2018 13th IAPR International Workshop on Document Analysis
  Systems (DAS)}, pages 43--48. IEEE.

\bibitem[Cobanoglu et~al., 2024]{Cobanoglu2024}
Cobanoglu, Y., S\'aenz, L., Khait, I., and Jim\'enez, E. (2024).
\newblock Sign detection for cuneiform tablets.
\newblock {\em it -- Information Technology}, (65).

\bibitem[Dencker et~al., 2020]{dencker2020deep}
Dencker, T., Klinkisch, P., Maul, S.~M., and Ommer, B. (2020).
\newblock Deep learning of cuneiform sign detection with weak supervision using
  transliteration alignment.
\newblock {\em Plos one}, 15(12):e0243039.

\bibitem[Devlin et~al., 2019]{devlin2019bertpretrainingdeepbidirectional}
Devlin, J., Chang, M.-W., Lee, K., and Toutanova, K. (2019).
\newblock Bert: Pre-training of deep bidirectional transformers for language
  understanding.

\bibitem[{electronic Babylonian Library (eBL)}, 2026]{eBLPlatform}
{electronic Babylonian Library (eBL)} (2026).
\newblock The ``electronic babylonian library'' (ebl) platform.
\newblock \url{https://www.ebl.lmu.de/}.
\newblock Database platform, accessed 2026-05-20.

\bibitem[Ester et~al., 1996]{ester1996dbscan}
Ester, M., Kriegel, H.-P., Sander, J., and Xu, X. (1996).
\newblock A density-based algorithm for discovering clusters in large spatial
  databases with noise.
\newblock In {\em Proceedings of the 2nd International Conference on Knowledge
  Discovery and Data Mining (KDD)}, pages 226--231. AAAI Press.

\bibitem[Garces~Arias et~al., 2023]{arias-etal-2023-automatic}
Garces~Arias, E., Pai, V., Sch{\"o}ffel, M., Heumann, C., and A{\ss}enmacher,
  M. (2023).
\newblock Automatic transcription of handwritten old {O}ccitan language.
\newblock In Bouamor, H., Pino, J., and Bali, K., editors, {\em Proceedings of
  the 2023 Conference on Empirical Methods in Natural Language Processing},
  pages 15416--15439, Singapore. Association for Computational Linguistics.

\bibitem[Gordin et~al., 2024]{gordin2024cured}
Gordin, S., Alper, M., Romach, A., Santos, L.~S., Yochai, N., and Lalazar, R.
  (2024).
\newblock Cured: Deep learning optical character recognition for cuneiform text
  editions and legacy materials.
\newblock In {\em Proceedings of the 1st Workshop on Machine Learning for
  Ancient Languages (ML4AL 2024)}, pages 130--140.

\bibitem[Kahle et~al., 2017]{8270253}
Kahle, P., Colutto, S., Hackl, G., and M\"uhlberger, G. (2017).
\newblock Transkribus---a service platform for transcription, recognition and
  retrieval of historical documents.
\newblock In {\em 2017 14th IAPR International Conference on Document Analysis
  and Recognition (ICDAR)}, volume~04, pages 19--24.

\bibitem[Koch et~al., 2023]{koch-etal-2023-tailored}
Koch, P., Nu{\~n}ez, G.~V., Garces~Arias, E., Heumann, C., Sch{\"o}ffel, M.,
  H{\"a}berlin, A., and Assenmacher, M. (2023).
\newblock A tailored handwritten-text-recognition system for medieval {L}atin.
\newblock In Anderson, A., Gordin, S., Li, B., Liu, Y., and Passarotti, M.~C.,
  editors, {\em Proceedings of the Ancient Language Processing Workshop}, pages
  103--110, Varna, Bulgaria. INCOMA Ltd., Shoumen, Bulgaria.

\bibitem[Lewenstein et~al., 2026]{lewenstein2026large}
Lewenstein, O., L\'opez, D., Dankwardt, C., Alrawi, M.~F., Grill, L., Mak, B.,
  Set\"al\"a, A., Gori, F., H\"atinen, A., Rauchhaus, F., F\"oldi, Z., and
  Jim\'enez, E. (2026).
\newblock A large-scale dataset of annotated cuneiform sign images for digital
  palaeography.
\newblock {\em Journal of Open Humanities Data}.

\bibitem[Li et~al., 2022]{li2022trocrtransformerbasedopticalcharacter}
Li, M., Lv, T., Chen, J., Cui, L., Lu, Y., Florencio, D., Zhang, C., Li, Z.,
  and Wei, F. (2022).
\newblock Trocr: Transformer-based optical character recognition with
  pre-trained models.

\bibitem[Lin et~al., 2014]{Lin2014COCO}
Lin, T.-Y., Maire, M., Belongie, S., Hays, J., Perona, P., Ramanan, D.,
  Doll{\'a}r, P., and Zitnick, C.~L. (2014).
\newblock Microsoft coco: Common objects in context.
\newblock In Fleet, D., Pajdla, T., Schiele, B., and Tuytelaars, T., editors,
  {\em Computer Vision -- ECCV 2014}, pages 740--755, Cham. Springer
  International Publishing.

\bibitem[Liu et~al., 2016]{liu2016ssd}
Liu, W., Anguelov, D., Erhan, D., Szegedy, C., Reed, S., Fu, C.-Y., and Berg,
  A.~C. (2016).
\newblock Ssd: Single shot multibox detector.
\newblock In {\em European conference on computer vision}, pages 21--37.
  Springer.

\bibitem[Liu and Jin, 2017]{liu2017deep}
Liu, Y. and Jin, L. (2017).
\newblock Deep matching prior network: Toward tighter multi-oriented text
  detection.
\newblock In {\em Proceedings of the IEEE conference on computer vision and
  pattern recognition}, pages 1962--1969.

\bibitem[Liu et~al., 2021]{liu2021swintransformerhierarchicalvision}
Liu, Z., Lin, Y., Cao, Y., Hu, H., Wei, Y., Zhang, Z., Lin, S., and Guo, B.
  (2021).
\newblock Swin transformer: Hierarchical vision transformer using shifted
  windows.

\bibitem[Mikulinsky et~al., 2025]{mikulinsky2025protosnap}
Mikulinsky, R., Alper, M., Gordin, S., Jim{\'e}nez, E., Cohen, Y., and
  Averbuch-Elor, H. (2025).
\newblock Protosnap: Prototype alignment for cuneiform signs.
\newblock In {\em Proceedings of the International Conference on Learning
  Representations (ICLR)}.

\bibitem[Pavlopoulos et~al., 2024]{pavlopoulos-etal-2024-challenging}
Pavlopoulos, J., Kougia, V., Garces~Arias, E., Platanou, P., Shabalin, S.,
  Liagkou, K., Papadatos, E., Essler, H., Camps, J.-B., and Fischer, F. (2024).
\newblock Challenging error correction in recognised byzantine {G}reek.
\newblock In Pavlopoulos, J., Sommerschield, T., Assael, Y., Gordin, S., Cho,
  K., Passarotti, M., Sprugnoli, R., Liu, Y., Li, B., and Anderson, A.,
  editors, {\em Proceedings of the 1st Workshop on Machine Learning for Ancient
  Languages (ML4AL 2024)}, pages 1--12, Hybrid in Bangkok, Thailand and online.
  Association for Computational Linguistics.

\bibitem[Radford et~al., 2019]{Radford2019LanguageMA}
Radford, A., Wu, J., Child, R., Luan, D., Amodei, D., and Sutskever, I. (2019).
\newblock Language models are unsupervised multitask learners.

\bibitem[Reade, 2017]{reade2017manufacture}
Reade, J.~E. (2017).
\newblock The manufacture, evaluation and conservation of clay tablets
  inscribed in cuneiform: Traditional problems and solutions.
\newblock {\em Iraq}, 79:163--202.

\bibitem[Saeed et~al., 2024]{saeed2024create}
Saeed, E.~A., Jasim, A.~D., and Malik, M. A.~A. (2024).
\newblock Create distinctive databases of ancient languages and using a
  computer vision model to accurately recognise and classify them.
\newblock {\em Data in Brief}, 56:110809.

\bibitem[Sarawgi et~al., 2025]{sarawgi2025digitizingnepalswrittenheritage}
Sarawgi, A., Arias, E.~G., and Zotter, C. (2025).
\newblock Digitising nepal's written heritage: A comprehensive htr pipeline for
  old nepali manuscripts.

\bibitem[Simonjetz et~al., 2024]{Simonjetz2024}
Simonjetz, F., Laasonen, J., Cobanoglu, Y., Fraser, A., and Jim\'enez, E.
  (2024).
\newblock Reconstruction of cuneiform literary texts as text matching.

\bibitem[Streck, 2010]{Streck2010}
Streck, M.~P. (2010).
\newblock Gro\ss{}es fach altorientalistik: Der umfang des keilschriftlichen
  textkorpus.
\newblock {\em Mitteilungen der Deutschen Orient-Gesellschaft}, (142):35--58.

\bibitem[Wang et~al., 2011]{wang2011end}
Wang, K., Babenko, B., and Belongie, S. (2011).
\newblock End-to-end scene text recognition.
\newblock In {\em 2011 International conference on computer vision}, pages
  1457--1464. IEEE.

\bibitem[Wang et~al., 2012]{wang2012end}
Wang, T., Wu, D.~J., Coates, A., and Ng, A.~Y. (2012).
\newblock End-to-end text recognition with convolutional neural networks.
\newblock In {\em Proceedings of the 21st international conference on pattern
  recognition (ICPR2012)}, pages 3304--3308. IEEE.

\bibitem[Wigington et~al., 2018]{wigington2018start}
Wigington, C., Tensmeyer, C., Davis, B., Barrett, W., Price, B., and Cohen, S.
  (2018).
\newblock Start, follow, read: End-to-end full-page handwriting recognition.
\newblock In {\em Proceedings of the European conference on computer vision
  (ECCV)}, pages 367--383.

\bibitem[Williams et~al., 2025]{williams2025deepscribe}
Williams, E.~C., Su, G., Schloen, S.~R., Prosser, M., Paulus, S., and Krishnan,
  S. (2025).
\newblock Deepscribe: localisation and classification of elamite cuneiform
  signs via deep learning.
\newblock {\em ACM Journal on Computing and Cultural Heritage}, 18(2):1--32.

\bibitem[Yesiltepe et~al., 2019]{Yesiltepe2019}
Yesiltepe, B., Asuroglu, T., and Aktas, A.~Z. (2019).
\newblock Computerised {H}ittite {C}uneiform {S}ign {R}ecognition and
  {K}nowledge-{B}ased {S}ystem {A}pplication examples.
\newblock {\em European Scientific Journal}, 15(33):32--53.

\bibitem[Zhou et~al., 2017]{zhou2017east}
Zhou, X., Yao, C., Wen, H., Wang, Y., Zhou, S., He, W., and Liang, J. (2017).
\newblock East: an efficient and accurate scene text detector.
\newblock In {\em Proceedings of the IEEE conference on Computer Vision and
  Pattern Recognition}, pages 5551--5560.

\bibitem[Zhu et~al., 2021a]{zhu2021deformabledetrdeformabletransformers}
Zhu, X., Su, W., Lu, L., Li, B., Wang, X., and Dai, J. (2021a).
\newblock Deformable detr: Deformable transformers for end-to-end object
  detection.

\bibitem[Zhu et~al., 2021b]{zhu2021fourier}
Zhu, Y., Chen, J., Liang, L., Kuang, Z., Jin, L., and Zhang, W. (2021b).
\newblock Fourier contour embedding for arbitrary-shaped text detection.
\newblock In {\em Proceedings of the IEEE/CVF conference on computer vision and
  pattern recognition}, pages 3123--3131.

\end{thebibliography}
\end{document}